\documentclass{article}
\usepackage[nonatbib,preprint]{neurips_2025}

\usepackage[utf8]{inputenc} % allow utf-8 input
\usepackage[T1]{fontenc}    % use 8-bit T1 fonts
\usepackage{multirow}
\usepackage{microtype}
\usepackage{amsmath}
\usepackage{amssymb}
\usepackage{booktabs}
\usepackage{graphicx}
\usepackage{enumitem}
\usepackage{subcaption}
\usepackage{xcolor}
\usepackage{stackengine}
\usepackage{tabularx}
\usepackage{balance}
\usepackage{placeins}
\usepackage[detect-all,separate-uncertainty = true]{siunitx}
\definecolor{citecolor}{rgb}{0.086,0.04,0.50}
\usepackage[breaklinks,colorlinks=true,citecolor=citecolor,linkcolor=citecolor]{hyperref}
\usepackage[capitalise]{cleveref}

\usepackage[backref=true,natbib=true]{biblatex}

\creflabelformat{equation}{#2#1#3}

\makeatletter
\renewcommand{\paragraph}{%
  \@startsection{paragraph}{4}%
  {\z@}{1.5ex \@plus 1ex \@minus .2ex}{-0.5em}%
  {\normalfont\normalsize\bfseries}%
}
\makeatother

\setlist{leftmargin=18pt,nosep}

\title{Interpretable and Testable Vision \\Features via Sparse Autoencoders}

\author{Samuel Stevens\\
The Ohio State University\\
{\tt\small stevens.994@osu.edu}
\And
Wei-Lun Chao\\
The Ohio State University\\
{\tt\small chao.209@osu.edu}
\AND
Tanya Berger-Wolf\\
The Ohio State University\\
{\tt\small berger-wolf.1@osu.edu}
\And
Yu Su\\
The Ohio State University\\
{\tt\small su.809@osu.edu}
}

\begin{document}

\maketitle

\begin{abstract}
To truly understand vision models, we must not only interpret their learned features but also validate these interpretations through controlled experiments. 
While earlier work offers either rich semantics or direct control, few post-hoc tools supply both in a single, model-agnostic procedure.
We use sparse autoencoders (SAEs) to bridge this gap; each sparse feature comes with real-image exemplars that reveal its meaning and a decoding vector that can be manipulated to probe its influence on downstream task behavior. 
By applying our method to widely-used pre-trained vision models, we reveal meaningful differences in the semantic abstractions learned by different pre-training objectives. 
We then show that a single SAE trained on frozen ViT activations supports patch-level causal edits across tasks (classification and segmentation) all without retraining the ViT or task heads.
These qualitative, falsifiable demonstrations position SAEs as a practical bridge between concept discovery and causal probing of vision models.
We provide code, demos and models on our project website: \url{https://osu-nlp-group.github.io/saev}.
% We will release code, pretrained SAE weights, and an interactive web demo to facilitate further work.
\end{abstract}

\section{Introduction}\label{sec:introduction}

Understanding deep neural networks requires more than passive observation; it demands the ability to test hypotheses through controlled intervention.
This mirrors the scientific method itself: true understanding emerges not from mere observation, but from our ability to make and test predictions through controlled experiments. 
Biologists did not truly understand how genes control traits until they could manipulate DNA and observe the effects. 
Similarly, understanding neural networks requires not just observation, but systematically testing explanations through controlled experiments.

Applying the scientific method to understanding vision models requires three key capabilities. 
First, we need observable features that correspond to human-interpretable concepts like textures, objects, or abstract properties. 
Biologists need measurable markers of gene expression; we need reliable ways to identify specific visual concepts within our models. 
Second, we must be able to precisely manipulate these features to test hypotheses about their causal role, like geneticists' knockout experiments to validate gene function. 
Finally, methods must work with existing models, just as biological techniques must work with existing organisms rather than requiring genetic redesign.

The scientific method of observe, hypothesize and intervene maps neatly onto machine learning interpretability: observe model behavior, hypothesize why a model makes a particular prediction, then perturb the system to see how behavior changes \citep{poincare1914science,popper1959logic}. 
Saliency and attribution maps \citep{simonyan2013deep,selvaraju2017gradcam}, feature-visualization \citep{mordvintsev2015deepdream,olah2017feature} and network-dissection \citep{bau2017broden} methods link model activations to human concepts, acting as a hypothesized explanation for model behavior.
Yet translating those hypotheses into controlled tests is fraught with challenges: masking-based ablations risk distribution shift \citep{hooker2019benchmark}; other surveys conclude that quantitative faithfulness tests remain elusive \citep{adebayo2018sanity,distill2020attribution}.
Concept activation vectors \citep{kim2018tcav,goyal2019counterfactual} translate human concepts into testable relationships, but require hand-labeled concept sets.
Concept-bottleneck models embed editable concepts directly in the architecture \citep{koh2020concept}, but doing so demands retraining the model. 
Consequently, few post-hoc techniques offer (i) human-interpretable features, (ii) a natural intervention on those features, and (iii) compatibility with pre-trained vision models. 

We show in this work that sparse autoencoders (SAEs) offer a natural solution to this problem.
SAEs transform dense, entangled activation vectors into higher-dimensional sparse representations in which each element likely corresponds to a distinct semantic concept.
Because the SAE is trained to reconstruct the original activations, each element also explicitly corresponds to a particular direction in the original dense activation space. 
Thus, each element in the sparse representation has both (1) exemplar patches, yielding a semantic visual description, and (2) a corresponding decoding vector in the original space, yielding a precise method for causal interventions.
For example, a blue-jay wing activates an element whose exemplar patches are nothing but blue feathers. 
Suppressing model activations in that direction pushes the classifier away from blue-feathered species, completing the intervene step and confirming the feature's causal role (see \cref{fig:classification} for an example).
In this work, we demonstrate that SAEs are a natural, intuitive post-hoc method for the complete observe–hypothesize–intervene cycle.

\begin{figure*}
    \centering
    \small
    \includegraphics[width=\textwidth]{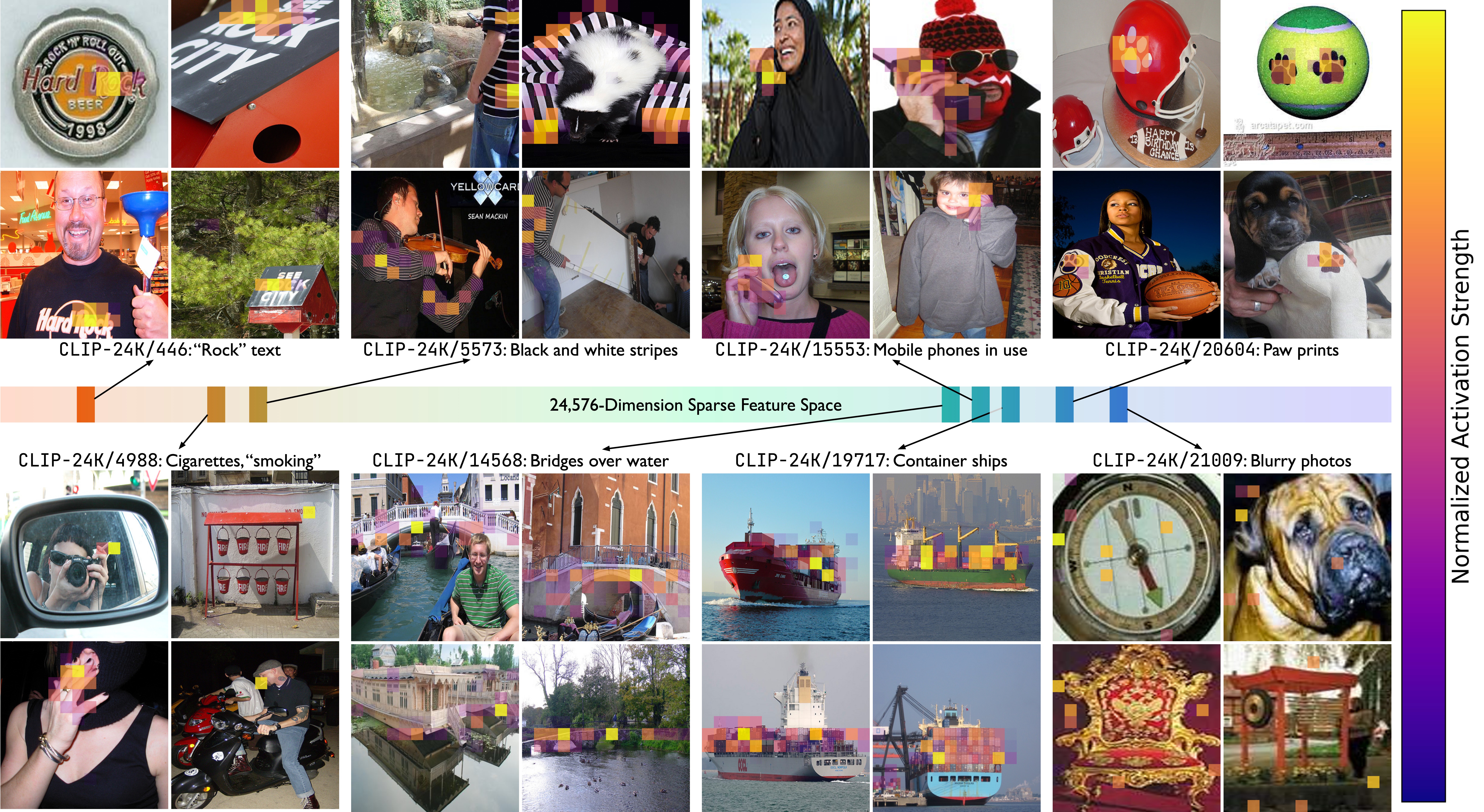}
    \caption{Sparse autoencoders (SAEs) trained on pre-trained ViT activations discover a wide spread of features across both visual patterns and semantic structures. 
    We show eight different features from an SAE trained on ImageNet-1K activations from a CLIP-trained ViT-B/16.
    Colored patches mark where the SAE features fire within an image; each SAE feature fires on on semantically consistent but visually diverse patches.
    }\label{fig:overview2}
    \vspace{-8pt}
\end{figure*}

First, we use SAEs to survey learned features in CLIP \citep{radford2021learning} and DINOv2 \citep{oquab2023dinov2} and illustrate consistent differences (\cref{sec:understanding}). 
We discover that CLIP learns to recognize country-specific visual patterns (\cref{sec:cultural-understanding}) and style-agnostic representations (\cref{sec:semantic-abstraction}), suggesting that language supervision leads to rich world knowledge that purely visual training cannot achieve.
Just as comparative biology reveals how different environments shape evolution \citep{grant2006evolution,losos2011lizards,brawand2014genomic}, our analysis shows that different training objectives lead to qualitatively different internal representations.

Second, we validate our SAE interpretations on two canonical tasks. 
When models detect features corresponding to bird markings, we confirm their causal role by showing that modifying them predictably changes species classification (\cref{sec:classification}). 
Similarly, with semantic segmentation, we show that identified features enable precise, targeted manipulation: we can suppress specific semantic concepts (like ``sand'' or ``grass'') while preserving all other scene elements, demonstrating both the semantic meaning of our features and their independence from unrelated concepts (\cref{sec:semseg}).
Interpretability is still, as \citeauthor{olah2024qualitative} put it, ``an early, messy, un-established science in which low-hanging qualitative structure is often more trustworthy than synthetic summary scores'' \citep{olah2024qualitative}. 
We therefore follow their advice, and the cautions of \citeauthor{adebayo2018sanity} and \citeauthor{hooker2019benchmark} by prioritizing human-falsifiable interventions over aggregate metrics for the present work.
Our public, interactive webapps enable readers to explore these features and perform their own causal edits, turning our claims into falsifiable, interactive evidence.\footnote{\href{https://anonymous-saev.github.io/saev/demos/semseg}{\url{https://anonymous-saev.github.io/saev/demos/semseg}} and \href{https://anonymous-saev.github.io/saev/demos/classification}{\url{https://anonymous-saev.github.io/saev/demos/classification}}}

Third, we provide a public, extensible codebase that works with any vision transformer, enabling broad investigation of modern vision models.
We demonstrate this flexibility through fine-grained classification and semantic segmentation experiments, with future updates planned.

Through these contributions, we demonstrate that sparse autoencoders are a practical method to link semantic feature discovery and interactive causal testing in modern vision models.
%, something previous SAE demos have not evaluated on downstream tasks.

\section{Related Work}\label{sec:related-work}

Saliency methods and attribute maps~\citep{zhou2016learning,sundararajan2017axiomatic, selvaraju2017gradcam,wang2020scorecam} highlight relevant parts of images for a given prediction; these methods excel at observation but require masking to test causality, which can introduce distribution shift \citep{adebayo2018sanity,hooker2019benchmark}.
Feature visualization methods~\citep{simonyan2013deep,zeiler2014visualizing,mordvintsev2015inceptionism,olah2017feature} reveal interpretable concepts through synthetic images. 
While these approaches demonstrate that models learn meaningful features, the generated images are unrealistic and we cannot validate if these visualizations actually drive model behavior. 
In contrast, SAEs identify features through real image examples and enables direct testing of their causal influence.
Network dissection~\citep{bau2017broden,zhou2018interpreting,hernandez2021natural,ghiasi2022vision} attempts to map individual neurons to semantic concepts using labeled datasets. 
However, this approach struggles when single neurons encode multiple concepts \citep{goh2021multimodal}. 
In contrast, SAEs (and dictionary learning in general) are explicitly designed to naturally decompose polysemantic representations into single-concept interpretable components.
Concept activation vectors \cite[CAVs,][]{kim2018tcav,goyal2019counterfactual,ghorbani2019towards,singla2021causal} provide a natural intervention once a concept dataset exists, but they depend on hand-labeled positive examples and deliver one vector per concept, limiting exploratory use.
We train SAEs on pre-trained models and explore the models' visual vocabularies without labeling any examples.

Concept bottleneck models (CBMs; \citealt{ghorbani2019towards,koh2020concept}), prototype-based approaches \citep{chen2019looks,nauta2021looks,donnelly2022deformable,willard2024looks}, and prompting-based methods
\citep{paul2024simple,chowdhury2025prompt} explicitly incorporate interpretable concepts into model architecture. 
While powerful, these methods require specific pre-training model architectures and objectives and cannot analyze existing pre-trained models. 
SAEs can be applied to and analyze any pre-trained vision transformer.
Follow-up work addresses these shortcomings: 
\citet{yuksekgonul2023posthoc} convert pre-trained models to CBMS; \citet{schrodi2024concept,tan2024explain} develop CBMs with open vocabularies rather than a fixed concept set.
In contrast to these works, SAEs reveal a model’s intrinsic knowledge in a task‐agnostic manner by decomposing dense activations into sparse, monosemantic features without any retraining. 
This plug‐and‐play approach not only faithfully captures the vision model’s internal representations but also enables precise interventions, making SAEs a more flexible tool for validating model interpretation.

Most similar to our approach is DN-CBM \citep{rao2024discover}, which trains sparse autoencoders on CLIP embeddings and then names units via text similarity before training a new linear concept bottleneck model.
Our work differs in three respects:
(i) we leave the backbone and task head frozen, editing activations directly;
(ii) we validate edits on both classification and semantic segmentation, demonstrating task agnosticism;
(iii) we analyze two pre-trained models (CLIP vs DINOv2) to study representation differences.

Sparse autoencoders are a popular dictionary learning method for extracting concept libraries.
\citet{makhzani2013k,makhzani2015winner} apply $k$-sparse autoencoders to learn improved image representations.
\citet{subramanian2018spine} apply $k$-sparse autoencoders to static word embeddings (word2vec \citep{mikolov2013word2vec} and GloVe \citep{pennington2014glove}) to improve interpretability.
\citet{zhang2019word,yun2021transformer,bricken2023monosemanticity,templeton2024scaling,gao2024scaling} apply sparse autoencoders to transformer-based language model activations and find highly interpretable features.
\citet{lim2025sparse} apply SAEs to vision models to explain prompt-based tuning methods and \citet{thasarathan2025universal} apply SAEs to vision models for cross-model concept alignment. 
We aim to convincingly demonstrate that feature descriptions (exemplar images) aligned with causal interventions (decoder vectors) are a natural fit for the observe-hypothesize-intervene loop.
% Several blog posts demonstrate SAEs trained on CLIP visual embeddings \cite{fry2024multimodal,daujotas2024interpreting,daujotas2024case}; 

% Through this synthesis of interpretation, control, and sparsity, our work enables both understanding and manipulation of vision model behavior in ways previously impossible. The SAE framework provides a unified approach to discovering interpretable features and validating their causal influence through controlled intervention.

\section{Methodology}\label{sec:methodology}

Vision models demonstrate remarkable capabilities, but understanding their behavior requires rigorous scientific investigation. 
We present a framework that enables systematic observation, hypothesis formation, and experimental validation using sparse autoencoders (SAEs).

\subsection{Observations: Feature Discovery}
Given a pre-trained vision model, we first need reliable ways to observe its internal representations. 
Traditional approaches like visualizing individual neurons or studying attention patterns provide limited insight. 
Instead, we train SAEs on activation vectors from intermediate model layers, enabling systematic observation of learned features.

\begin{figure*}
    \centering
    \small
    \includegraphics[width=\textwidth]{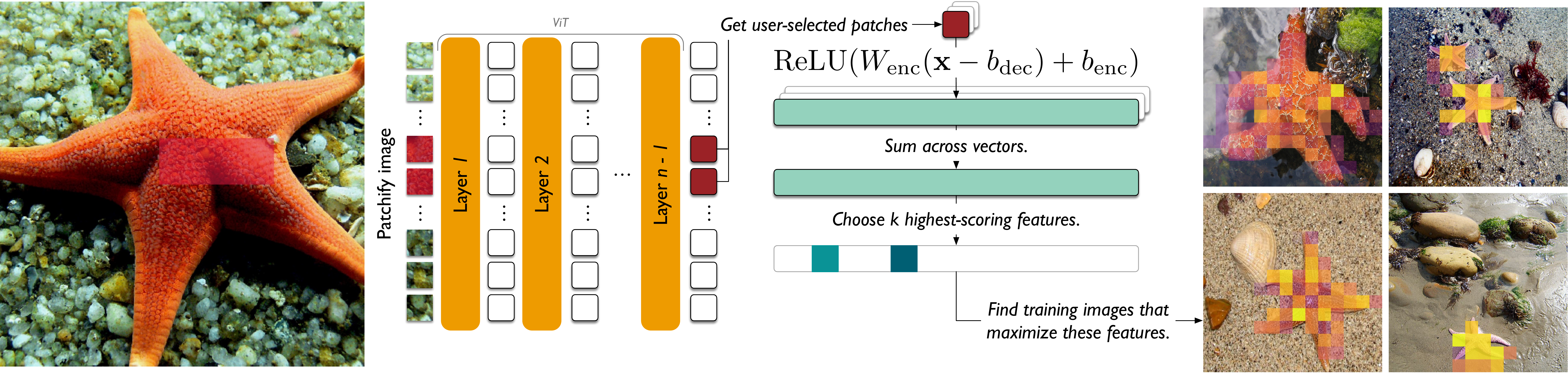}
    \caption{Given a picture and a set of highlighted patches, we find exemplar images by (1) getting ViT activations for each patch, (2) computing a sparse representation for each highlighted patch (\cref{eq:enc,eq:act}), (3) summing over sparse representations, (4) choosing the top $k$ features by activation magnitude and (5) finding existing images that maximize these features.}\label{fig:overview}
    \vspace{-8pt}
\end{figure*}

\subsection{Hypotheses: SAE-Generated Explanations}\label{sec:sae-training}

Sparse autoencoders generate testable hypotheses by decomposing dense activation vectors into sparse feature vectors. 
Given an $d$-dimensional activation vector $\mathbf{x} \in \mathbb{R}^d$ from an intermediate layer $l$ of a vision transformer, an SAE maps $\mathbf{x}$ to a sparse representation $f(\mathbf{x})$ (\cref{eq:enc,eq:act}) and reconstructs the original input (\cref{eq:dec}). 
We use ReLU autoencoders \citep{bricken2023monosemanticity,templeton2024scaling}:
\begin{align}
    \mathbf{h} &= W_\text{enc} (\mathbf{x} - b_\text{dec}) + b_\text{enc} \label{eq:enc} \\
    f(\mathbf{x}) &= \text{ReLU}(\mathbf{h}) \label{eq:act} \\
    \mathbf{\hat{x}} &= W_\text{dec} f(\mathbf{x}) + b_\text{dec} \label{eq:dec}
\end{align}
where $W_\text{enc} \in \mathbb{R}^{n \times d}$, $b_\text{enc} \in \mathbb{R}^{n}$, $W_\text{dec} \in \mathbb{R}^{d \times n}$ and $b_\text{dec} \in \mathbb{R}^{d}$.
The training objective minimizes reconstruction error while encouraging sparsity:
\begin{equation}
    \mathcal{L}(\theta) = ||\mathbf{x} - \mathbf{\hat{x}}||_2^2 + \lambda \mathcal{S}(f(\mathbf{x}))
\end{equation}
where $\lambda$ controls the sparsity penalty and $\mathcal{S}$ measures sparsity (L1 norm for training, L0 for model selection).

We train on $N$ randomly sampled patch activation vectors from the residual stream of layer $l$ in a vision transformer. 
Following prior work \citep{templeton2024scaling}, we subtract the mean activation vector and normalize activation vectors to unit norm before training.
% Detailed reproduction instructions are \href{https://osu-nlp-group.github.io/SAE-V/saev/#guide-to-training-saes-on-vision-models}{available}.
See \cref{app:training-details} for additional details.
Given a pre-trained ViT like CLIP or DINOv2, an image, and one or more patches of interest, we leverage a trained SAE to find similar examples (\cref{fig:overview}).

\subsection{Experiments: Testing Through Control}

We validate SAE-proposed explanations through a general intervention framework that leverages the common pipeline of vision tasks: an image is first converted into $p$ $d$-dimensional activation vectors in $\mathbb{R}^{p \times d}$ (e.g., from a vision transformer), then these activation vectors are mapped by a task-specific decoder to produce outputs in the task-specific output space $\mathcal{O}$. 
For instance, in semantic segmentation each patch's activation vector $\mathbb{R}^d$ is fed to a decoder head that assigns pixel-level segmentation labels, and these pixel labels are assembled into the final segmentation map.

Given a task-specific decoder $\mathcal{M}\colon \mathbb{R}^{p \times d} \rightarrow \mathcal{O}$ that maps from $n$ activations vectors to per-patch class predictions, our intervention process proceeds in six steps:
\begin{enumerate}
   \item{Encode and reconstruct: $f(\mathbf{x}) = \text{ReLU}(W_\text{enc} (\mathbf{x} - b_\text{dec}) + b_\text{enc})$ and $\mathbf{\hat{x}} = W_\text{dec} f(\mathbf{x}) + b_\text{dec}$.}
   \item{Calculate reconstruction error \citep{templeton2024scaling}: $\mathbf{e} = \mathbf{x} - \mathbf{\hat{x}}$.}
   \item{Modify individual values of $f(\mathbf{x})$ to get $f(\mathbf{x})'$.}
   \item{Reconstruct modified activations: $\mathbf{\hat{x}}' = W_\text{dec} f(\mathbf{x})' + b_\text{dec}$.}
   \item{Add back error: $\mathbf{x}' = \mathbf{e} + \mathbf{\hat{x}}'$.}
   \item{Compare outputs $\mathcal{M}(\mathbf{x})$ versus $\mathcal{M}(\mathbf{x}')$.}
\end{enumerate}
In plain language: We start by converting an image into a set of activation vectors (one per patch) that capture the ViT's internal representation.
Then, we encode these vectors into a sparse representation that highlights the key features, while also tracking the small differences (errors) between the sparse form and the original. 
Next, we deliberately tweak the sparse representation to modify a feature of interest. 
After reconstructing the modified activation vectors (and adding back the previously captured details), we pass them through the task-specific decoder (e.g., for classification, segmentation, or another vision task) to see how the output changes. 
By comparing the original and altered outputs, we can determine whether and how the targeted feature influences the model's behavior.

% Feature modifications are scaled relative to training-time maximum activations, with values in $[-20, 20]$ representing $-20\times$ to $20\times$ the maximum observed activation. 
% To identify relevant features for modification, we combine automated metrics (like F1 scores between feature activations and ground truth) with manual verification through activation visualization (\cref{sec:control-experiments}).

\begin{figure*}[t]
    \centering
    \small
    \setlength{\tabcolsep}{1pt}
    \begin{tabular}{ccccccc}
        \multicolumn{4}{c}{(a) \texttt{CLIP-24K/6909}: ``Brazil''} & \multicolumn{2}{c}{(b) \textit{Not} Brazil} & \multirow{3}{*}{\includegraphics[width=19pt]{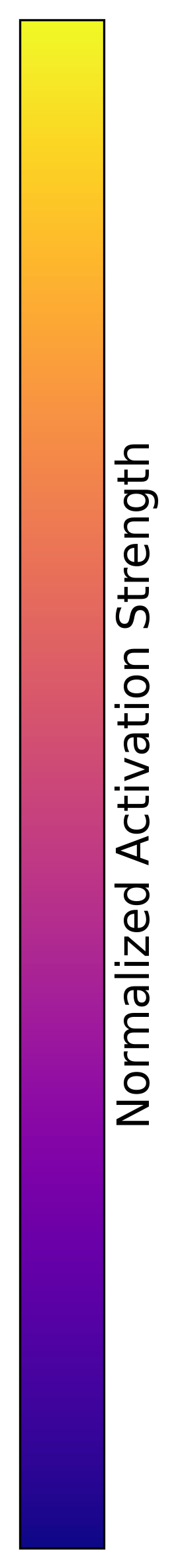}} \\
        \cmidrule(lr){1-4} \cmidrule(lr){5-6}
        \includegraphics[width=0.15\textwidth]{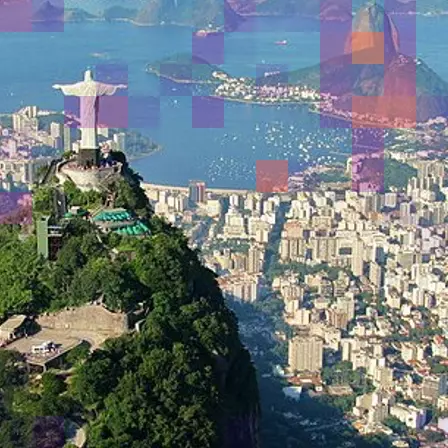} & 
        \includegraphics[width=0.15\textwidth]{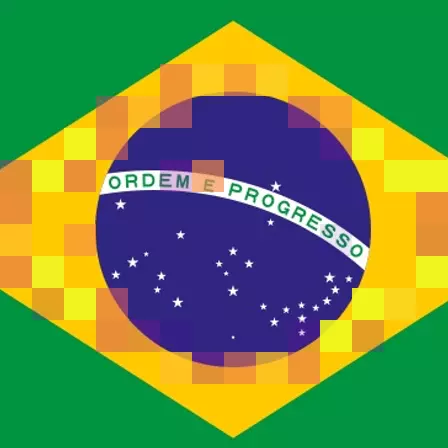} & 
        \includegraphics[width=0.15\textwidth]{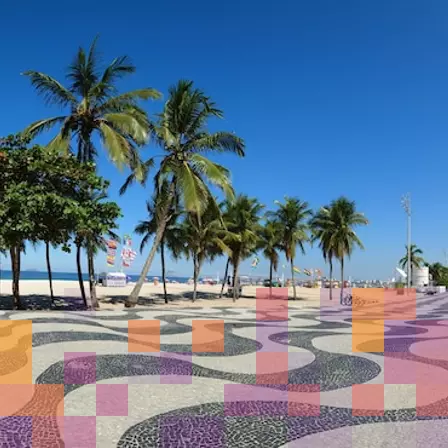} & 
        \includegraphics[width=0.15\textwidth]{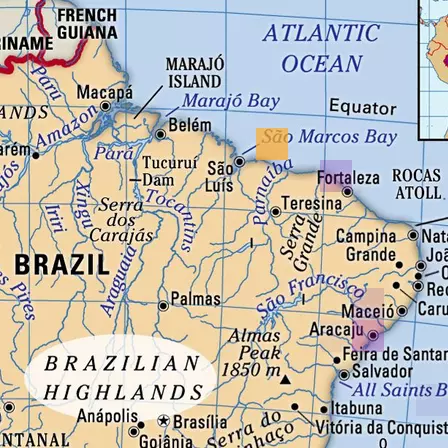} &
        \includegraphics[width=0.15\textwidth]{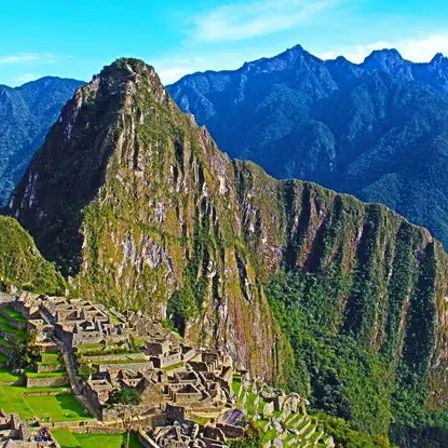} &
        \includegraphics[width=0.15\textwidth]{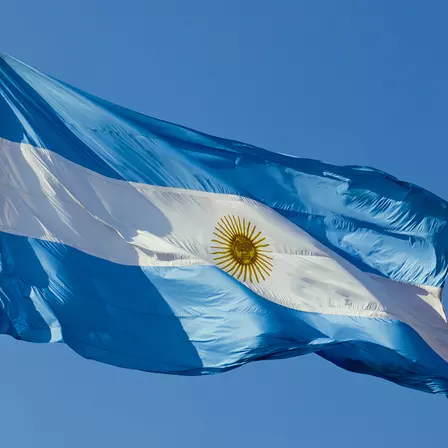} \\[12pt]
        \multicolumn{4}{c}{(c) \texttt{DINOv2-24K/9823}} & \multicolumn{2}{c}{(d) ImageNet-1K Exemplars} \\ 
        \cmidrule(lr){1-4} \cmidrule(lr){5-6}
        \includegraphics[width=0.15\textwidth]{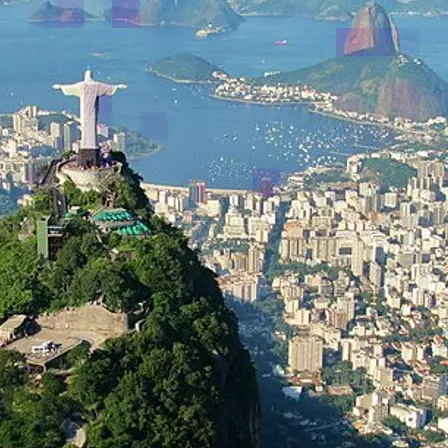} & 
        \includegraphics[width=0.15\textwidth]{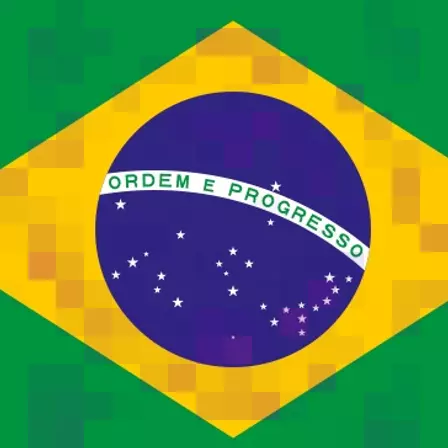} & 
        \includegraphics[width=0.15\textwidth]{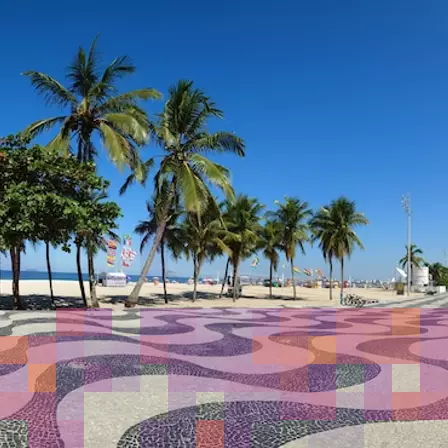} & 
        \includegraphics[width=0.15\textwidth]{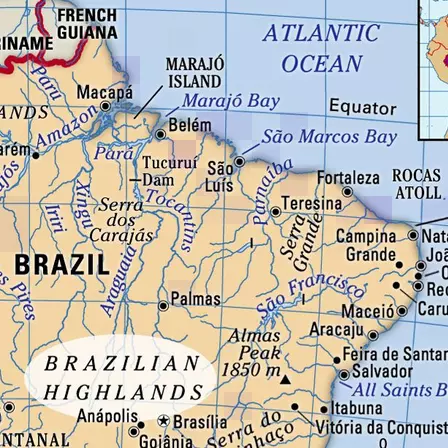} &
        \includegraphics[width=0.15\textwidth]{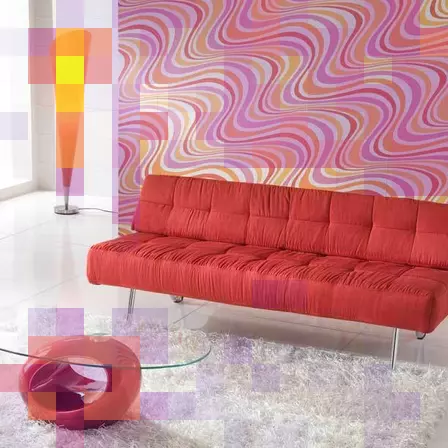} &
        \includegraphics[width=0.15\textwidth]{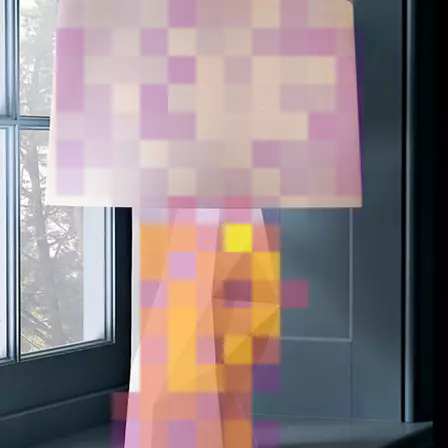} 
    \end{tabular}
    \caption{
        CLIP learns robust cultural visual features.
        \textbf{Top Left (a):} A ``Brazil'' feature (\texttt{CLIP-24K/6909}) responds to distinctive Brazilian imagery including Rio de Janeiro's urban landscape, the national flag, and the iconic sidewalk tile pattern of Copacabana Beach
        \textbf{Top Right (b):} \texttt{CLIP-24K/6909} does not respond to other South American symbols like Machu Picchu or the Argentinian flag.
        \textbf{Bottom Left (c):} We search DINOv2's SAE for a similar ``Brazil'' feature and find that \texttt{DINOv2-24K/9823} fires on Brazilian imagery.
        \textbf{Bottom Right (d):} However, maximally activating ImageNet-1K examples for \texttt{DINOv2-24K/9823} are of lamps, convincing us that \texttt{DINOv2-24K/9823} does not reliably detect Brazilian cultural symbols.
    }\label{fig:clip-vs-dinov2-cultural}
    \vspace{-8pt}
\end{figure*}

\section{SAE-Enabled Analysis of Vision Models}\label{sec:understanding}

Sparse autoencoders (SAEs) provide a useful new lens for understanding and comparing vision models. 
By decomposing dense activation vectors into interpretable features, SAEs enable systematic analysis of what different architectures learn and how their training objectives shape their internal representations. 
We demonstrate that even simple manual inspection of top-activating images for SAE-discovered features can reveal differences between models that would be difficult to detect through other methods.

While prior work has used techniques like TCAV \citep{kim2018tcav} to probe for semantic concepts in vision models, these methods typically test for pre-defined concepts and rely on supervised concept datasets. 
In contrast, our SAE-based approach discovers interpretable features directly from model activations without requiring concept supervision. 
We train SAEs on intermediate layer activations following the procedure detailed in \cref{sec:methodology}, then analyze the highest-activating image patches for each learned feature as in \cref{fig:overview}. 
This straightforward process reveals both specific features (e.g., a feature that fires on dental imagery) and model-level patterns (e.g., one model consistently learning more abstract features than another).
Technical details of our process are provided in \cref{app:understanding-details}.

We refer to individual SAE features using \texttt{MODEL-WIDTH/INDEX}, where MODEL identifies the vision transformer the SAE was trained on (e.g., CLIP or DINOv2), WIDTH indicates the number of features in the SAE (e.g., 24K for \num{24576}), and INDEX uniquely identifies the specific feature. 
For example, \texttt{CLIP-24K/20652} refers to feature \num{20652} from a \num{24576}-dimensional SAE trained on CLIP activations.

\begin{figure*}[t]
\centering
\begin{minipage}[b]{0.46\textwidth}
\centering
\small
\captionof{table}{SAE metrics on ImageNet-1K. Low reconstruction error (mean-squared error; MSE) and sparse activations demonstrate successful decomposition of ViT representations. ``Dead'' neurons are active on less than $10^{-7}\%$ of inputs; ``Dense'' neurons are active on more than $1\%$ of all inputs.
}\label{tab:training-metrics}
\begin{tabular}{lrrrr}
\toprule
Model & MSE & L0 & Dead & Dense \\
\midrule
CLIP & 0.0761 & 412.7 & 0 & 11,858 \\
DINOv2 & 0.0697 & 728.7 & 1 & 19,735 \\
\bottomrule
\end{tabular}
\end{minipage}
\hfill 
\begin{minipage}[t]{0.46\textwidth} 
\centering
\small
\includegraphics[width=\textwidth]{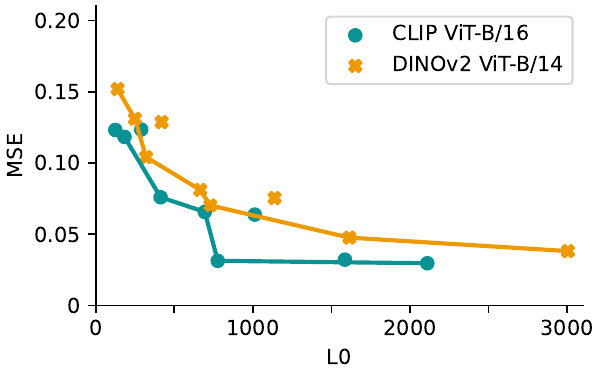}
\vspace{-18pt}
\caption{MSE and L0 on the training dataset for all learning rates and sparsity coefficents $\lambda$.}
\label{fig:training-metrics}
\end{minipage}
\vspace{-12pt}
\end{figure*}

\subsection{Language Enables Cultural Understanding}\label{sec:cultural-understanding}

We analyze SAE features and find that CLIP learns remarkably robust representations of cultural and geographic concepts--a capability that appears absent in DINOv2's purely visual representations. 
This demonstrates how language supervision enables learning abstract cultural features that persist across diverse visual manifestations.

We find individual SAE features that consistently activate on imagery associated with specific countries, while remaining inactive on visually similar but culturally distinct images (\cref{fig:clip-vs-dinov2-cultural}). 
For instance, CLIP reliably detects German visual elements across architectural landmarks (like the Brandenburg Gate), sports imagery (German national team uniforms), and cultural symbols (Oktoberfest celebrations).
Crucially, this feature remains inactive on other European architectural landmarks or sporting events, suggesting it captures genuine cultural associations rather than just visual similarities.

Similarly, we find features that activate on distinctly Brazilian imagery, spanning Rio de Janeiro's urban landscape, the national flag, coastal scenes, and cultural celebrations. 
These features show selective activation, responding strongly to Brazilian content while remaining inactive on visually similar scenes from other South American locations. 
This selective response pattern suggests CLIP has learned to recognize and group culturally related visual elements, even when they share few low-level visual features.
See \cref{fig:additional-clip-vs-dinov2-cultural} for additional examples of this phenomena.

In contrast, when we analyze DINOv2's features, we find no comparable country-specific representations. 
This difference illustrates how language supervision could help CLIP to learn culturally meaningful visual abstractions that pure visual training does not discover.

\begin{figure*}[t]
    \centering
    \small
    \setlength{\tabcolsep}{1pt}
    \begin{tabular}{ccccccc}
        \multicolumn{4}{c}{\texttt{CLIP-24K/20652}} & \multicolumn{2}{c}{ImageNet-1K Exemplars} & \multirow{3}{*}{\includegraphics[width=19pt]{figures/clip-vs-dinov2/legend.png}} \\ 
        \cmidrule(lr){1-4} \cmidrule(lr){5-6}
        \includegraphics[width=0.15\textwidth]{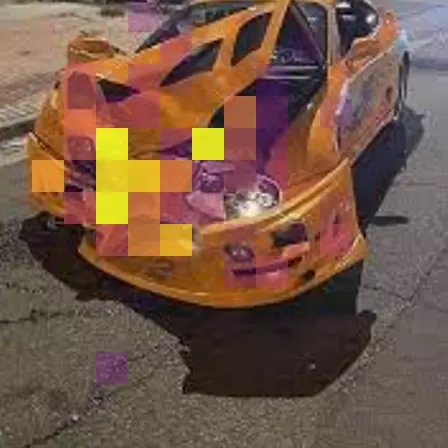} & 
        \includegraphics[width=0.15\textwidth]{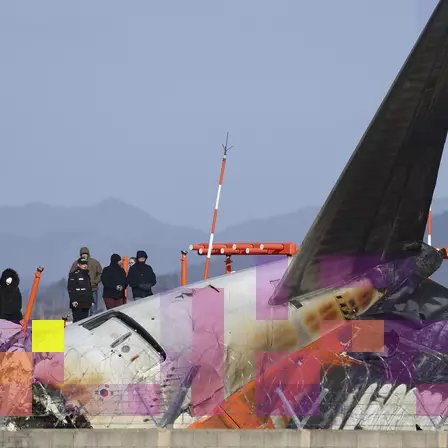} & 
        \includegraphics[width=0.15\textwidth]{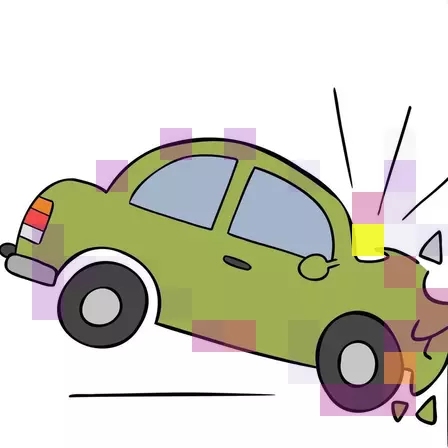} &
        \includegraphics[width=0.15\textwidth]{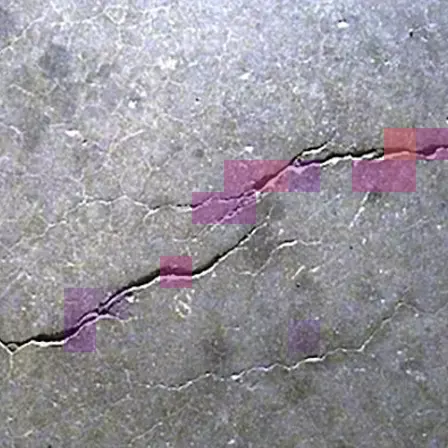} &
        \includegraphics[width=0.15\textwidth]{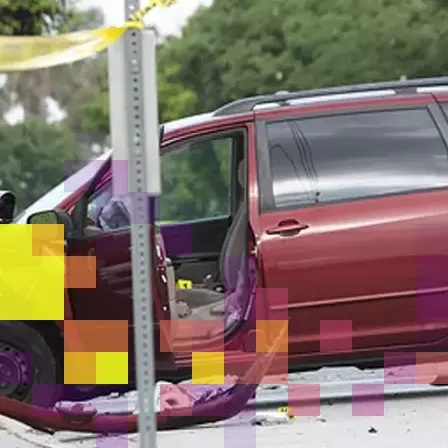} &
        \includegraphics[width=0.15\textwidth]{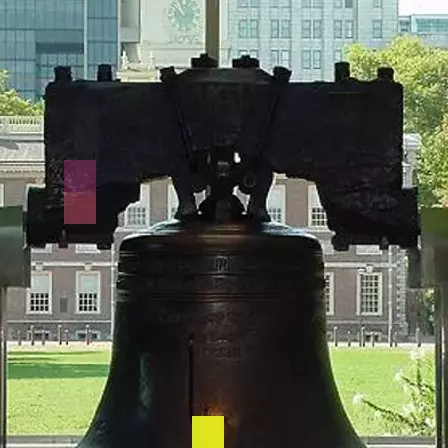} \\[12pt]
        \multicolumn{4}{c}{\texttt{DINOv2-24K/9672}} & \multicolumn{2}{c}{ImageNet-1K Exemplars} \\ 
        \cmidrule(lr){1-4} \cmidrule(lr){5-6}
        \includegraphics[width=0.15\textwidth]{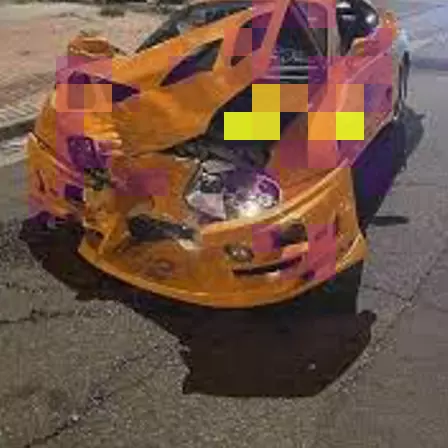} & 
        \includegraphics[width=0.15\textwidth]{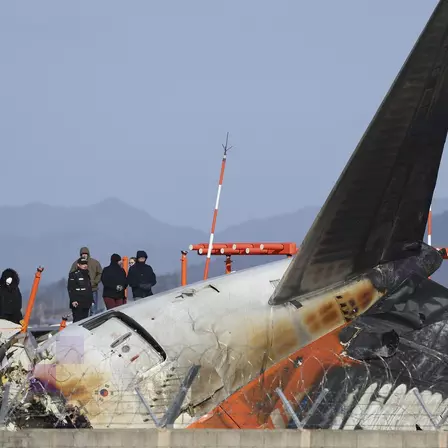} & 
        \includegraphics[width=0.15\textwidth]{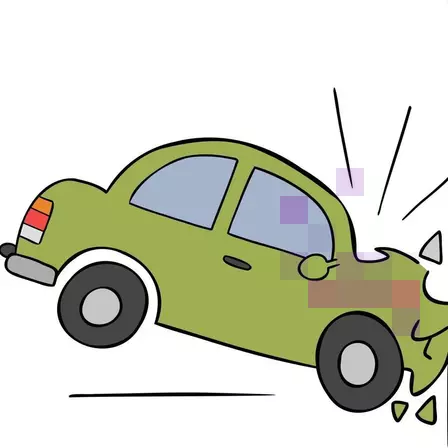} &
        \includegraphics[width=0.15\textwidth]{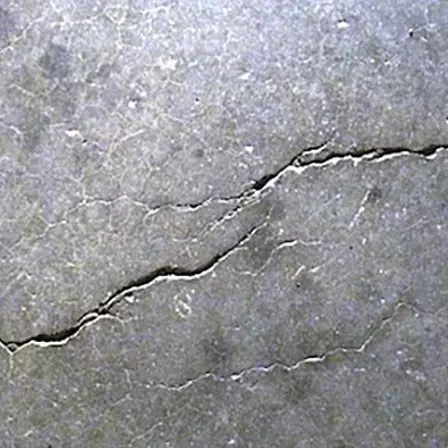} &
        \includegraphics[width=0.15\textwidth]{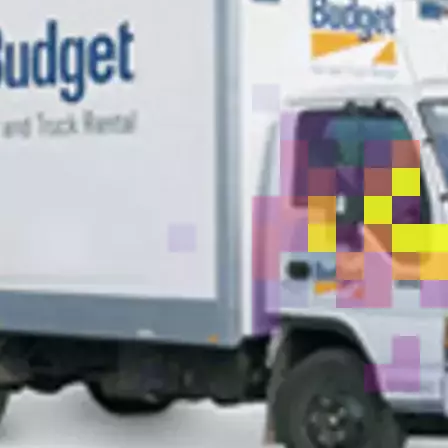} &
        \includegraphics[width=0.15\textwidth]{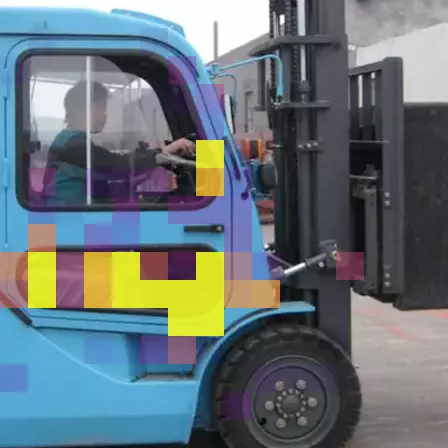}\\
    \end{tabular}
    \caption{
        CLIP learns unified representations of abstract concepts that persist across visual styles.
        Highlighted patches indicate feature activation strength.
        \textbf{Upper Left:} We find a CLIP SAE feature (\texttt{CLIP-24K/20652}) that consistently activates on ``accidents'' or ``crashes'': car accidents, plane crashes, cartoon depictions of crashes and generally damaged metal.
        \textbf{Upper Right:} Two exemplar images from ImageNet-1K for feature \texttt{CLIP-24K/20652}.
        \textbf{Lower Left:} We probe an SAE trained on DINOv2 activations. \texttt{DINOv2-24K/9762} is the closest feature, but does not reliably fire on all the examples.
        \textbf{Lower Right:} Two exemplar images from ImageNet-1K for feature \texttt{DINOv2-24K/9762} clarifies that it does not match the semantic concept of ``crash.''
    }\label{fig:clip-vs-dinov2-semantic}
    \vspace{-8pt}
\end{figure*}

\subsection{Language Induces Semantic Abstraction}\label{sec:semantic-abstraction}

Beyond cultural concepts, CLIP learns abstract semantic features that persist across visual styles, a capability DINOv2 lacks, revealing how language supervision enables human-like semantic learning.

A striking example emerges in representations of mechanical accidents and crashes. 
We discover a feature (\texttt{CLIP-24K/20652}) that activates on accident scenes across photographs of car accidents, crashed planes and cartoon depictions of crashes (\cref{fig:clip-vs-dinov2-semantic}). 
This feature fires on both a news photograph of a car accident or a stylized illustration of a collision--suggesting CLIP learns an abstract representation of ``accident'' that transcends visual style.

In contrast, DINOv2's features fragment these examples across multiple low-level visual patterns. 
While it learns features that detect specific visual aspects of accidents (like crumpled metal in photographs), no single feature captures the semantic concept across different visual styles. 
This suggests that without language supervision, DINOv2 lacks the learning signal needed to unite these diverse visual presentations into a single abstract concept.
% This pattern extends beyond accident scenes to other semantic concepts, from object states to event types (see \cref{app:additional-clip-vs-dinov2-semantic} for additional examples). 
We hypothesize that CLIP's language supervision provides explicit signals to group visually distinct but semantically related images, while DINOv2's purely visual training offers no such bridge across visual styles. 
The ability to form such abstract semantic concepts, independent of visual style, is a meaningful difference in how these models process and represent visual information.

\subsection{Implications for Vision Model Selection}

\citet{tong2024eyes} constructed a challenging dataset for vision-language models (VLMs; \citealt{liu2024llava,liu2024llava1_5,lu2024deepseek_vl}) using only CLIP as a visual encoder.
Prior work \citep{jiang2023clip,shi2024eagle,tong2024cambrian,li2025eagle} demonstrates empirical benefits from combining different vision encoders.
However, the mechanistic basis for these improvements has remained unclear.

Our SAE-driven analysis provides a possible explanation for these empirical findings.
The distinct feature patterns we observe--CLIP's abstract semantic concepts versus DINOv2's style-specific features--suggest these models develop complementary rather than redundant internal representations.
When CLIP learns to recognize ``accidents'' across various visual styles, or country-specific features across diverse contexts, it develops abstractions that may help with high-level semantic understanding. 
Meanwhile, DINOv2's more granular features could provide detailed visual information that complements CLIP's abstract representations.
Rather than treating encoders as interchangeable components to be evaluated purely on benchmark performance, practitioners might choose encoders based on characterized feature patterns.

% However, further work is needed to fully understand how these distinct representational strategies contribute to model performance across different tasks and domains.

\begin{figure*}[t]
    \centering
    \small
    \includegraphics[width=\linewidth]{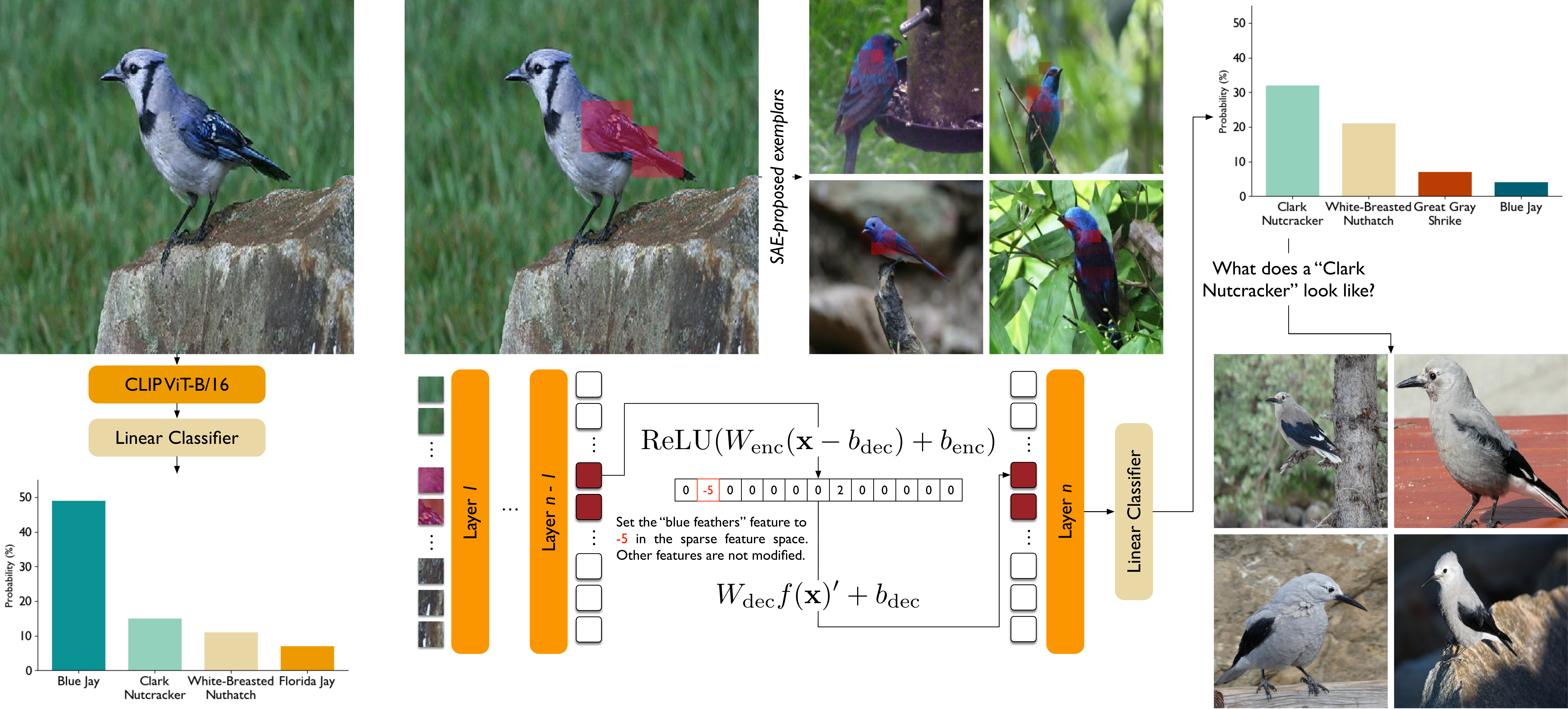}
    \caption{
    Demonstrating the scientific method for understanding vision model behavior using sparse autoencoders (SAEs). 
    \textbf{Left:} We observe that CLIP predicts ``Blue Jay.'' 
    \textbf{Upper Middle:} We select the bird's wing in the input image; the SAE proposes a hypothesis that the most salient feature is ``blue feathers'' via exemplar images.
    \textbf{Lower Middle:} We validate this hypothesis through controlled intervention by suppressing the identified ``blue feathers'' feature in the model's activation space.
    \textbf{Right:} we observe a change in behavior: the predicted class shifts away from ``Blue Jay'' towards ``Clark Nutcracker'', a similar bird besides the lack of blue plumage. 
    This three-step process of observation, hypothesis formation, and validation enables systematic investigation of how vision models process visual information.
    % \todo{(1) use plasma colormap for the examples.}
    }\label{fig:classification}
    \vspace{-8pt}
\end{figure*}

\section{Testing Hypotheses of Vision Model Behavior}\label{sec:control-experiments}

Understanding how vision models process information requires testing that our explanations accurately capture the model's behavior. 
While prior work has demonstrated interpretable features or model control in isolation, validating explanations requires showing both that discovered features are meaningful and that we can precisely manipulate them. 
We demonstrate that SAE-derived features enable reliable control across two complementary visual understanding tasks--classification and segmentation--providing strong evidence that our interpretations faithfully capture model behavior.

% Vision models process information across multiple levels of abstraction, from local visual features to global semantic reasoning. 
% Standard evaluation approaches focus on a single level of processing, limiting their ability to validate feature explanations. 
% We overcome this by strategically selecting tasks that test different aspects of visual understanding: classification validates precise control over visual attributes and segmentation demonstrates spatial coherence of identified features,
% and visual question answering proves our manipulations preserve rather than destroy learned semantic representations.

For each task, we first train or utilize an existing task-specific prediction head. 
We then perform controlled feature interventions using our SAE and compare model behavior before and after intervention. 
Success across these diverse challenges, which we demonstrate through diverse qualitative results, provides strong evidence that our method identifies meaningful and faithful features.

\subsection{Image Classification}\label{sec:classification}

Precisely manipulating individual visual features is essential for validating our understanding of vision model decisions. 
While prior work identifies features through post-hoc analysis, testing causal relationships requires demonstrating that controlled feature manipulation yields predictable output changes.
Using fine-grained bird classification as a testbed, we show that SAE-derived features enable precise control: we can manipulate specific visual attributes like beak shape or plumage color while preserving other traits, producing predictable changes in classification outputs that validate our feature-level explanations of model behavior.

We train a sparse autoencoder on activations from a CLIP-pretrained ViT-B/16 \citep{radford2021learning}, using the complete ImageNet-1K dataset \citep{russakovsky2015imagenet} to ensure broad feature coverage.
We record activations from layer \num{11} (of \num{12} total layers); we argue that the second-to-last layer contains features that are highly semantic but not overfit to CLIP's contrastive image-text matching objective.
The SAE uses a $32\times$ expansion factor (\num{24576} features) to capture a rich vocabulary of visual concepts. 
Through an interactive interface, we identify interpretable features by examining patches that maximally activate specific SAE features. 
For example, selecting the blue feathers of a blue jay (\textit{Cyanocitta cristata}) reveals SAE features that consistently activate on similar colorations across the dataset.

To validate the proposed feature explanation, we can manipulate them by adjusting their values in the SAE's latent space.
When we reduce the ``blue feathers''  feature in a blue jay image, the model's prediction shifts from blue jay to Clark's nutcracker (\textit{Nucifraga columbiana})--a semantically meaningful change that aligns with ornithological knowledge (\cref{fig:classification}).
See \cref{app:classification-details} for SAE details, linear classifier details, and additional qualitative examples.
While our current interface limits manipulation to single features at a time, the approach generalizes beyond bird classification and provides a useful new tool for understanding vision model decisions. 
% Some manipulations can lead to unexpected results, highlighting areas for future investigation into feature interactions and stability.

\begin{figure*}[t]
    \centering
    \small
    \includegraphics[width=\linewidth]{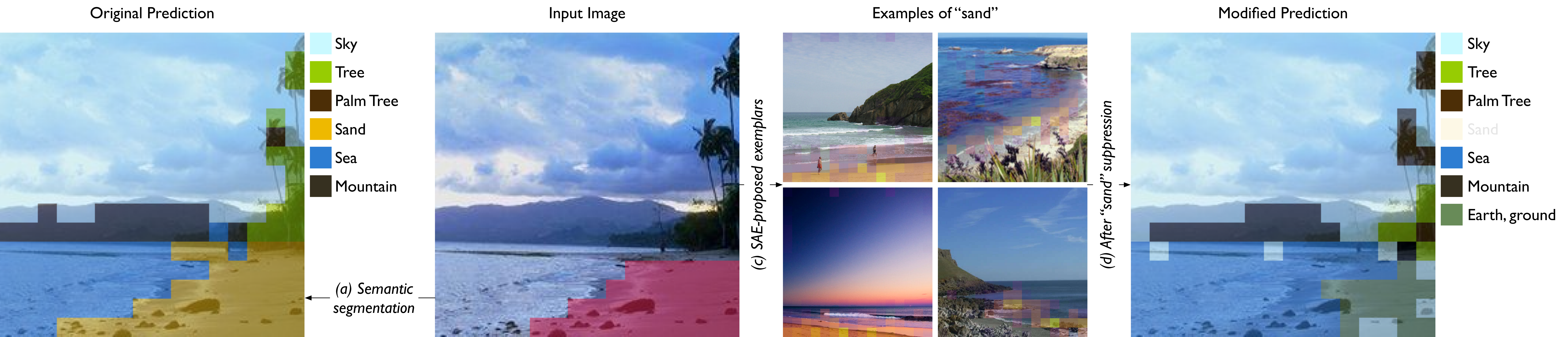}
    \caption{\textbf{Far Left:} We train a linear head to predict semantic segmentation classes for each patch.
    \textbf{Middle Left:} We choose all ``sand'' patches in the input image to inspect. 
    \textbf{Middle Right:} Our SAE proposes exemplar images for the maximally activating sparse dimension, as in \cref{sec:classification}, suggesting that DINOv2 learns a sand feature.
    \textbf{Far Right:} We suppress the sand feature in not just the selected patches, but \textit{all} patches. 
    We modify all activation vectors before DINOv2's final transformer layer followed by our trained linear segmentation head. We see that the head predicts ``earth, ground'' and ``water'' for the former sand patches. Both classes are good second choices if ``sand'' is unavailable. Notably, other patches are not meaningfully affected, demonstrating the pseudo-orthogonality of the SAE's learned feature vectors.
    }\label{fig:semseg}
    \vspace{-8pt}
\end{figure*}

\subsection{Semantic Segmentation}\label{sec:semseg}

When discovered features are entangled, attempts to modify one aspect of an image often produce unintended changes in others, making precise intervention impossible.
Through semantic segmentation, we observe that SAE features behave as a pseudo-orthogonal basis for the model's representational space; while not mathematically perpendicular, these features are, in practice, functionally independent.
When we suppress semantic concepts like ``sand'' or ``grass'' \textit{in all patches}, we observe consistent changes in targeted regions while leaving other predictions intact, demonstrating this functional independence.

We train a linear probe on vision model patch-level features on the ADE20K semantic segmentation dataset \citep{zhou2017ade20k} following the setup in \citet{oquab2023dinov2}.
Specifically, given an image-level representation tensor of $p \times p$ patch vectors of dimension $d$ ($\mathbb{R}^{p \times p \times d}$), we train a linear probe to predict a low-resolution $p \times p \times C$ logit map, which is up-sampled to full resolution.
We achieve \num{35.1} mIoU on the ADE20K validation set.
While this method falls short of state-of-the-art methods, it is a sufficient testbed for studying feature control.
We train a sparse autoencoder on ImageNet-1K activations from layer 11 of a DINOv2-pretrained ViT-B/16 following the procedure described in \cref{sec:methodology}.
We switch from CLIP to DINOv2 to demonstrate that our experimental validation approach generalizes across model families with distinct training objectives and because DINOv2 provides richer spatial representations for dense prediction tasks like semantic segmentation.
Details are are available in \cref{app:semseg-details}.

To validate our method's ability to manipulate semantic features, we developed an interactive interface that allows precise control over individual SAE features. 
Users can select specific image patches and modify their feature activations in both positive and negative directions, enabling targeted manipulation of semantic concepts. 
This approach provides several advantages over automated evaluation metrics: it allows exploration of feature interactions, demonstrates the spatial coherence of manipulations, and reveals how features compose to form higher-level concepts. 
The interface makes it possible to test hypotheses about learned features through direct experimentation; for example, we can verify that ``sand'' features truly capture sand-like textures by suppressing them and observing consistent changes across diverse images, as shown in \cref{fig:semseg}.

\section{Conclusion}\label{sec:conclusion}

We demonstrate that sparse autoencoders enable both hypothesis formation and controlled testing for vision models, enabling empirically validated model interpretation.

Our work reveals fundamental insights about representation learning in vision models.
Our qualitative study suggests that language supervision helps CLIP develop abstractions that generalize across visual styles, a pattern absent in our observations of DINOv2.
This finding suggests that achieving human-like visual abstraction may require bridging between different modalities or forms of supervision.

We find that SAE-discovered features reliably capture genuine causal relationships in model behavior rather than mere correlations in classification and semantic segmentation. 
The ability to reliably manipulate these features while maintaining semantic coherence demonstrates that vision transformers learn decomposable, interpretable representations even without explicit supervision.

However, significant challenges remain. 
Current methods for identifying manipulable features still require manual exploration, and the relationship between feature interventions and model behavior becomes increasingly complex for higher-level tasks (see \cref{app:limitations} for additional discussion of limitations). 
Future work should focus on automating feature discovery, understanding feature interactions across different model architectures, and extending reliable control to more sophisticated visual reasoning tasks.

Just as scientific understanding emerges from our ability to both explain and manipulate natural phenomena, meaningful understanding of neural networks requires tools that unite explanation with experimentation. 
We believe SAEs provide a promising foundation for building such tools.

\clearpage
{
    \small
    \printbibliography

@book{poincare1914science,
title = {Science and Method},
author = {Poincar{\'e}, Henri},
year = 1914,
publisher = {Thomas Nelson},
address = {London}
}

@book{popper1959logic,
title = {The Logic of Scientific Discovery},
author = {Popper, Karl},
year = 1959,
publisher = {Julius Springer, Hutchinson \& Co},
address = {Berlin}
}

@inproceedings{pennington2014glove,
title = {Glove: Global vectors for word representation},
author = {Pennington, Jeffrey and Socher, Richard and Manning, Christopher D},
year = 2014,
booktitle = {Proceedings of the 2014 conference on empirical methods in natural language processing (EMNLP)},
pages = {1532--1543}
}

@article{mikolov2013word2vec,
title = {Distributed representations of words and phrases and their compositionality},
author = {Mikolov, Tomas and Sutskever, Ilya and Chen, Kai and Corrado, Greg S and Dean, Jeff},
year = 2013,
journal = {Advances in neural information processing systems},
volume = 26
}

@article{makhzani2013k,
title = {K-sparse autoencoders},
author = {Makhzani, Alireza and Frey, Brendan},
year = 2013,
journal = {arXiv preprint arXiv:1312.5663}
}

@article{makhzani2015winner,
title = {Winner-take-all autoencoders},
author = {Makhzani, Alireza and Frey, Brendan J},
year = 2015,
journal = {Advances in neural information processing systems},
volume = 28
}

@inproceedings{subramanian2018spine,
title = {Spine: Sparse interpretable neural embeddings},
author = {Subramanian, Anant and Pruthi, Danish and Jhamtani, Harsh and Berg-Kirkpatrick, Taylor and Hovy, Eduard},
year = 2018,
booktitle = {Proceedings of the AAAI conference on artificial intelligence},
volume = 32
}

@article{zhang2019word,
title = {Word embedding visualization via dictionary learning},
author = {Zhang, Juexiao and Chen, Yubei and Cheung, Brian and Olshausen, Bruno A},
year = 2019,
journal = {arXiv preprint arXiv:1910.03833}
}

@article{yun2021transformer,
title = {Transformer visualization via dictionary learning: contextualized embedding as a linear superposition of transformer factors},
author = {Yun, Zeyu and Chen, Yubei and Olshausen, Bruno A and LeCun, Yann},
year = 2021,
journal = {arXiv preprint arXiv:2103.15949}
}

@article{gao2024scaling,
title = {Scaling and evaluating sparse autoencoders},
author = {Gao, Leo and la Tour, Tom Dupr{\'e} and Tillman, Henk and Goh, Gabriel and Troll, Rajan and Radford, Alec and Sutskever, Ilya and Leike, Jan and Wu, Jeffrey},
year = 2024,
journal = {arXiv preprint arXiv:2406.04093}
}

@article{chowdhury2025prompt,
title = {Prompt-CAM: A Simpler Interpretable Transformer for Fine-Grained Analysis},
author = {Chowdhury, Arpita and Paul, Dipanjyoti and Mai, Zheda and Gu, Jianyang and Zhang, Ziheng and Mehrab, Kazi Sajeed and Campolongo, Elizabeth G and Rubenstein, Daniel and Stewart, Charles V and Karpatne, Anuj and others},
year = 2025,
journal = {arXiv preprint arXiv:2501.09333}
}

@inproceedings{paul2024simple,
title = {A simple interpretable transformer for fine-grained image classification and analysis},
author = {Paul, Dipanjyoti and Chowdhury, Arpita and Xiong, Xinqi and Chang, Feng-Ju and Carlyn, David and Stevens, Samuel and Provost, Kaiya L and Karpatne, Anuj and Carstens, Bryan and Rubenstein, Daniel and others},
year = 2024,
booktitle = {International Conference on Learning Representations}
}

@article{zhou2018interpreting,
title = {Interpreting deep visual representations via network dissection},
author = {Zhou, Bolei and Bau, David and Oliva, Aude and Torralba, Antonio},
year = 2018,
journal = {IEEE transactions on pattern analysis and machine intelligence},
publisher = {Ieee},
volume = 41,
number = 9,
pages = {2131--2145}
}

@inproceedings{bau2017broden,
title = {Network Dissection: Quantifying Interpretability of Deep Visual Representations},
author = {Bau, David and Zhou, Bolei and Khosla, Aditya and Oliva, Aude and Torralba, Antonio},
year = 2017,
booktitle = {2017 IEEE Conference on Computer Vision and Pattern Recognition (CVPR)},
volume = {},
number = {},
pages = {3319--3327},
doi = {10.1109/cvpr.2017.354}
}

@inproceedings{hernandez2021natural,
title = {Natural language descriptions of deep visual features},
author = {Hernandez, Evan and Schwettmann, Sarah and Bau, David and Bagashvili, Teona and Torralba, Antonio and Andreas, Jacob},
year = 2021,
booktitle = {International Conference on Learning Representations}
}

@article{bricken2023monosemanticity,
title = {Towards Monosemanticity: Decomposing Language Models With Dictionary Learning},
author = {Bricken, Trenton and Templeton, Adly and Batson, Joshua and Chen, Brian and Jermyn, Adam and Conerly, Tom and Turner, Nick and Anil, Cem and Denison, Carson and Askell, Amanda and Lasenby, Robert and Wu, Yifan and Kravec, Shauna and Schiefer, Nicholas and Maxwell, Tim and Joseph, Nicholas and Hatfield-Dodds, Zac and Tamkin, Alex and Nguyen, Karina and McLean, Brayden and Burke, Josiah E and Hume, Tristan and Carter, Shan and Henighan, Tom and Olah, Christopher},
year = 2023,
journal = {Transformer Circuits Thread},
note = {https://transformer-circuits.pub/2023/monosemantic-features/index.html}
}

@article{templeton2024scaling,
title = {Scaling Monosemanticity: Extracting Interpretable Features from Claude 3 Sonnet},
author = {Templeton, Adly and Conerly, Tom and Marcus, Jonathan and Lindsey, Jack and Bricken, Trenton and Chen, Brian and Pearce, Adam and Citro, Craig and Ameisen, Emmanuel and Jones, Andy and Cunningham, Hoagy and Turner, Nicholas L and McDougall, Callum and MacDiarmid, Monte and Freeman, C. Daniel and Sumers, Theodore R. and Rees, Edward and Batson, Joshua and Jermyn, Adam and Carter, Shan and Olah, Chris and Henighan, Tom},
year = 2024,
journal = {Transformer Circuits Thread},
url = {https://transformer-circuits.pub/2024/scaling-monosemanticity/index.html}
}

@inproceedings{radford2021learning,
title = {Learning transferable visual models from natural language supervision},
author = {Radford, Alec and Kim, Jong Wook and Hallacy, Chris and Ramesh, Aditya and Goh, Gabriel and Agarwal, Sandhini and Sastry, Girish and Askell, Amanda and Mishkin, Pamela and Clark, Jack and others},
year = 2021,
booktitle = {International conference on machine learning},
pages = {8748--8763},
organization = {Pmlr}
}

@misc{oquab2023dinov2,
title = {DINOv2: Learning Robust Visual Features without Supervision},
author = {Oquab, Maxime and Darcet, Timoth\'{e}e and Moutakanni, Theo and Vo, Huy V. and Szafraniec, Marc and Khalidov, Vasil and Fernandez, Pierre and Haziza, Daniel and Massa, Francisco and El-Nouby, Alaaeldin and Howes, Russell and Huang, Po-Yao and Xu, Hu and Sharma, Vasu and Li, Shang-Wen and Galuba, Wojciech and Rabbat, Mike and Assran, Mido and Ballas, Nicolas and Synnaeve, Gabriel and Misra, Ishan and Jegou, Herve and Mairal, Julien and Labatut, Patrick and Joulin, Armand and Bojanowski, Piotr},
year = 2023,
journal = {arXiv:2304.07193}
}

@misc{loshchilov2019adamw,
title = {Decoupled Weight Decay Regularization},
author = {Ilya Loshchilov and Frank Hutter},
year = 2019,
url = {https://arxiv.org/abs/1711.05101},
eprint = {1711.05101},
archiveprefix = {arXiv},
primaryclass = {cs.LG}
}

@article{ghiasi2022vision,
title = {What do vision transformers learn? a visual exploration},
author = {Ghiasi, Amin and Kazemi, Hamid and Borgnia, Eitan and Reich, Steven and Shu, Manli and Goldblum, Micah and Wilson, Andrew Gordon and Goldstein, Tom},
year = 2022,
journal = {arXiv preprint arXiv:2212.06727}
}

@article{templeton2024update,
title = {Update on Dictionary Learning Improvements},
author = {Templeton, Adly and Conerly, Tom and Marcus, Jonathan and Henighan, Tom},
year = 2024,
journal = {Transformer Circuits Thread},
url = {https://transformer-circuits.pub/2024/march-update/index.html\#dl-update}
}

@inproceedings{zhou2017ade20k,
title = {Scene parsing through ade20k dataset},
author = {Zhou, Bolei and Zhao, Hang and Puig, Xavier and Fidler, Sanja and Barriuso, Adela and Torralba, Antonio},
year = 2017,
booktitle = {Proceedings of the IEEE conference on computer vision and pattern recognition},
pages = {633--641}
}

@inproceedings{
lim2025sparse,
title={Sparse autoencoders reveal selective remapping of visual concepts during adaptation},
author={Hyesu Lim and Jinho Choi and Jaegul Choo and Steffen Schneider},
booktitle={The Thirteenth International Conference on Learning Representations},
year={2025},
url={https://openreview.net/forum?id=imT03YXlG2}
}

@misc{thasarathan2025universal,
title = {Universal Sparse Autoencoders: Interpretable Cross-Model Concept Alignment},
author = {Harrish Thasarathan and Julian Forsyth and Thomas Fel and Matthew Kowal and Konstantinos Derpanis},
year = 2025,
url = {https://arxiv.org/abs/2502.03714},
eprint = {2502.03714},
archiveprefix = {arXiv},
primaryclass = {cs.CV}
}

@article{fel2025archetypal,
title={Archetypal sae: Adaptive and stable dictionary learning for concept extraction in large vision models},
author={Fel, Thomas and Lubana, Ekdeep Singh and Prince, Jacob S and Kowal, Matthew and Boutin, Victor and Papadimitriou, Isabel and Wang, Binxu and Wattenberg, Martin and Ba, Demba and Konkle, Talia},
journal={arXiv preprint arXiv:2502.12892},
year={2025}
}

@article{distill2020attribution,
author = {Sturmfels, Pascal and Lundberg, Scott and Lee, Su-In},
title = {Visualizing the Impact of Feature Attribution Baselines},
journal = {Distill},
year = {2020},
note = {https://distill.pub/2020/attribution-baselines},
doi = {10.23915/distill.00022}
}

@misc{olah2024qualitative,
title = {Reflections on Qualitative Research},
author = {Chris Olah and Adam Jermyn},
year = 2024,
note = {Transformer Circuits blog},
howpublished = {\url{https://transformer-circuits.pub/2024/qualitative-essay/}}
}

@article{simonyan2013deep,
title = {Deep inside convolutional networks: Visualising image classification models and saliency maps},
author = {Simonyan, Karen},
year = 2013,
journal = {arXiv preprint arXiv:1312.6034}
}

@inproceedings{zeiler2014visualizing,
title = {Visualizing and Understanding Convolutional Networks},
author = {Zeiler, MD},
year = 2014,
booktitle = {European conference on computer vision/arXiv},
volume = 1311
}

@article{mordvintsev2015inceptionism,
title = {Inceptionism: Going deeper into neural networks},
author = {Mordvintsev, Alexander and Olah, Christopher and Tyka, Mike},
year = 2015,
journal = {Google research blog},
volume = 20,
number = 14,
pages = 5
}

@article{olah2017feature,
title = {Feature Visualization},
author = {Olah, Chris and Mordvintsev, Alexander and Schubert, Ludwig},
year = 2017,
journal = {Distill},
doi = {10.23915/distill.00007},
note = {https://distill.pub/2017/feature-visualization}
}

@article{mordvintsev2015deepdream,
title = {Deepdream-a code example for visualizing neural networks},
author = {Mordvintsev, Alexander and Olah, Christopher and Tyka, Mike},
year = 2015,
journal = {Google Research},
volume = 2,
number = 5
}

@inproceedings{zhou2016learning,
title = {Learning deep features for discriminative localization},
author = {Zhou, Bolei and Khosla, Aditya and Lapedriza, Agata and Oliva, Aude and Torralba, Antonio},
year = 2016,
booktitle = {Proceedings of the IEEE conference on computer vision and pattern recognition},
pages = {2921--2929}
}

@inproceedings{selvaraju2017gradcam,
title = {Grad-cam: Visual explanations from deep networks via gradient-based localization},
author = {Selvaraju, Ramprasaath R and Cogswell, Michael and Das, Abhishek and Vedantam, Ramakrishna and Parikh, Devi and Batra, Dhruv},
year = 2017,
booktitle = {Proceedings of the IEEE international conference on computer vision},
pages = {618--626}
}

@inproceedings{wang2020scorecam,
title={Score-CAM: Score-weighted visual explanations for convolutional neural networks},
author={Wang, Haofan and Wang, Zifan and Du, Mengnan and Yang, Fan and Zhang, Zijian and Ding, Sirui and Mardziel, Piotr and Hu, Xia},
booktitle={Proceedings of the IEEE/CVF conference on computer vision and pattern recognition workshops},
pages={24--25},
year={2020}
}

@inproceedings{sundararajan2017axiomatic,
title = {Axiomatic attribution for deep networks},
author = {Sundararajan, Mukund and Taly, Ankur and Yan, Qiqi},
year = 2017,
booktitle = {International conference on machine learning},
pages = {3319--3328},
organization = {Pmlr}
}

@inproceedings{kim2018tcav,
title = {Interpretability beyond feature attribution: Quantitative testing with concept activation vectors (tcav)},
author = {Kim, Been and Wattenberg, Martin and Gilmer, Justin and Cai, Carrie and Wexler, James and Viegas, Fernanda and others},
year = 2018,
booktitle = {International conference on machine learning},
pages = {2668--2677},
organization = {Pmlr}
}

@inproceedings{singla2021causal,
title={Using causal analysis for conceptual deep learning explanation},
author={Singla, Sumedha and Wallace, Stephen and Triantafillou, Sofia and Batmanghelich, Kayhan},
booktitle={Medical Image Computing and Computer Assisted Intervention--MICCAI 2021: 24th International Conference, Strasbourg, France, September 27--October 1, 2021, Proceedings, Part III 24},
pages={519--528},
year={2021},
organization={Springer}
}

@inproceedings{koh2020concept,
title = {Concept Bottleneck Models},
author = {Koh, Pang Wei and Nguyen, Thao and Tang, Yew Siang and Mussmann, Stephen and Pierson, Emma and Kim, Been and Liang, Percy},
year = 2020,
month = {7},
booktitle = {Proceedings of the 37th International Conference on Machine Learning},
publisher = {Pmlr},
series = {Proceedings of Machine Learning Research},
volume = 119,
pages = {5338--5348},
url = {https://proceedings.mlr.press/v119/koh20a.html},
editor = {III, Hal Daum\'{e} and Singh, Aarti}
}

@inproceedings{yuksekgonul2023posthoc,
title = {Post-hoc Concept Bottleneck Models},
author = {Mert Yuksekgonul and Maggie Wang and James Zou},
year = 2023,
booktitle = {The Eleventh International Conference on Learning Representations},
url = {https://openreview.net/forum?id=nA5AZ8CEyow}
}

@article{schrodi2024concept,
title = {Concept Bottleneck Models Without Predefined Concepts},
author = {Schrodi, Simon and Schur, Julian and Argus, Max and Brox, Thomas},
year = 2024,
journal = {arXiv preprint arXiv:2407.03921}
}

@inproceedings{tan2024explain,
title = {Explain via any concept: Concept bottleneck model with open vocabulary concepts},
author = {Tan, Andong and Zhou, Fengtao and Chen, Hao},
year = 2024,
booktitle = {European Conference on Computer Vision},
pages = {123--138},
organization = {Springer}
}

@inproceedings{rao2024discover,
title={Discover-then-name: Task-agnostic concept bottlenecks via automated concept discovery},
author={Rao, Sukrut and Mahajan, Sweta and B{\"o}hle, Moritz and Schiele, Bernt},
booktitle={European Conference on Computer Vision},
pages={444--461},
year={2024},
organization={Springer}
}

@inproceedings{goyal2019counterfactual,
title = {Counterfactual Visual Explanations},
author = {Goyal, Yash and Wu, Ziyan and Ernst, Jan and Batra, Dhruv and Parikh, Devi and Lee, Stefan},
year = 2019,
month = {6},
booktitle = {Proceedings of the 36th International Conference on Machine Learning},
publisher = {Pmlr},
series = {Proceedings of Machine Learning Research},
volume = 97,
pages = {2376--2384},
url = {https://proceedings.mlr.press/v97/goyal19a.html},
editor = {Chaudhuri, Kamalika and Salakhutdinov, Ruslan}
}

@inproceedings{ghorbani2019towards,
title = {Towards Automatic Concept-based Explanations},
author = {Ghorbani, Amirata and Wexler, James and Zou, James Y and Kim, Been},
year = 2019,
booktitle = {Advances in Neural Information Processing Systems},
publisher = {Curran Associates, Inc.},
volume = 32,
pages = {},
url = {https://proceedings.neurips.cc/paper_files/paper/2019/file/77d2afcb31f6493e350fca61764efb9a-Paper.pdf},
editor = {H. Wallach and H. Larochelle and A. Beygelzimer and F. d\textquotesingle Alch\'{e}-Buc and E. Fox and R. Garnett}
}

@article{goh2021multimodal,
author = {Goh, Gabriel and †, Nick Cammarata and †, Chelsea Voss and Carter, Shan and Petrov, Michael and Schubert, Ludwig and Radford, Alec and Olah, Chris},
title = {Multimodal Neurons in Artificial Neural Networks},
journal = {Distill},
year = {2021},
note = {https://distill.pub/2021/multimodal-neurons},
doi = {10.23915/distill.00030}
}

@article{chen2019looks,
title = {This looks like that: deep learning for interpretable image recognition},
author = {Chen, Chaofan and Li, Oscar and Tao, Daniel and Barnett, Alina and Rudin, Cynthia and Su, Jonathan K},
year = 2019,
journal = {Advances in neural information processing systems},
volume = 32
}

@inbook{nauta2021looks,
title = {This Looks Like That, Because ... Explaining Prototypes for Interpretable Image Recognition},
author = {Nauta, Meike and Jutte, Annemarie and Provoost, Jesper and Seifert, Christin},
year = 2021,
booktitle = {Machine Learning and Principles and Practice of Knowledge Discovery in Databases},
publisher = {Springer International Publishing},
pages = {441–456},
doi = {10.1007/978-3-030-93736-2\_34},
isbn = 9783030937362,
issn = {1865-0937},
url = {http://dx.doi.org/10.1007/978-3-030-93736-2_34}
}

@inproceedings{donnelly2022deformable,
title = {Deformable ProtoPNet: An Interpretable Image Classifier Using Deformable Prototypes},
author = {Donnelly, Jon and Barnett, Alina Jade and Chen, Chaofan},
year = 2022,
month = {6},
booktitle = {Proceedings of the IEEE/CVF Conference on Computer Vision and Pattern Recognition (CVPR)},
pages = {10265--10275}
}

@article{willard2024looks,
title = {This looks better than that: Better interpretable models with protopnext},
author = {Willard, Frank and Moffett, Luke and Mokel, Emmanuel and Donnelly, Jon and Guo, Stark and Yang, Julia and Kim, Giyoung and Barnett, Alina Jade and Rudin, Cynthia},
year = 2024,
journal = {arXiv preprint arXiv:2406.14675}
}

@article{adebayo2018sanity,
title = {Sanity checks for saliency maps},
author = {Adebayo, Julius and Gilmer, Justin and Muelly, Michael and Goodfellow, Ian and Hardt, Moritz and Kim, Been},
year = 2018,
journal = {Advances in neural information processing systems},
volume = 31
}

@article{hooker2019benchmark,
title = {A benchmark for interpretability methods in deep neural networks},
author = {Hooker, Sara and Erhan, Dumitru and Kindermans, Pieter-Jan and Kim, Been},
year = 2019,
journal = {Advances in neural information processing systems},
volume = 32
}

@article{tong2024cambrian,
title = {Cambrian-1: A fully open, vision-centric exploration of multimodal llms},
author = {Tong, Shengbang and Brown, Ellis and Wu, Penghao and Woo, Sanghyun and Middepogu, Manoj and Akula, Sai Charitha and Yang, Jihan and Yang, Shusheng and Iyer, Adithya and Pan, Xichen and others},
year = 2024,
journal = {arXiv preprint arXiv:2406.16860}
}

@inproceedings{tong2024eyes,
title = {Eyes wide shut? exploring the visual shortcomings of multimodal llms},
author = {Tong, Shengbang and Liu, Zhuang and Zhai, Yuexiang and Ma, Yi and LeCun, Yann and Xie, Saining},
year = 2024,
booktitle = {Proceedings of the IEEE/CVF Conference on Computer Vision and Pattern Recognition},
pages = {9568--9578}
}

@article{jiang2023clip,
title = {From clip to dino: Visual encoders shout in multi-modal large language models},
author = {Jiang, Dongsheng and Liu, Yuchen and Liu, Songlin and Zhao, Jin'e and Zhang, Hao and Gao, Zhen and Zhang, Xiaopeng and Li, Jin and Xiong, Hongkai},
year = 2023,
journal = {arXiv preprint arXiv:2310.08825}
}

@inproceedings{liu2024llava1_5,
title = {Improved baselines with visual instruction tuning},
author = {Liu, Haotian and Li, Chunyuan and Li, Yuheng and Lee, Yong Jae},
year = 2024,
booktitle = {Proceedings of the IEEE/CVF Conference on Computer Vision and Pattern Recognition},
pages = {26296--26306}
}

@article{liu2024llava,
title = {Visual instruction tuning},
author = {Liu, Haotian and Li, Chunyuan and Wu, Qingyang and Lee, Yong Jae},
year = 2024,
journal = {Advances in neural information processing systems},
volume = 36
}

@article{lu2024deepseek_vl,
title = {Deepseek-vl: towards real-world vision-language understanding},
author = {Lu, Haoyu and Liu, Wen and Zhang, Bo and Wang, Bingxuan and Dong, Kai and Liu, Bo and Sun, Jingxiang and Ren, Tongzheng and Li, Zhuoshu and Yang, Hao and others},
year = 2024,
journal = {arXiv preprint arXiv:2403.05525}
}

@article{li2025eagle,
title = {Eagle 2: Building Post-Training Data Strategies from Scratch for Frontier Vision-Language Models},
author = {Li, Zhiqi and Chen, Guo and Liu, Shilong and Wang, Shihao and VS, Vibashan and Ji, Yishen and Lan, Shiyi and Zhang, Hao and Zhao, Yilin and Radhakrishnan, Subhashree and others},
year = 2025,
journal = {arXiv preprint arXiv:2501.14818}
}

@article{shi2024eagle,
title = {Eagle: Exploring the design space for multimodal llms with mixture of encoders},
author = {Shi, Min and Liu, Fuxiao and Wang, Shihao and Liao, Shijia and Radhakrishnan, Subhashree and Huang, De-An and Yin, Hongxu and Sapra, Karan and Yacoob, Yaser and Shi, Humphrey and others},
year = 2024,
journal = {arXiv preprint arXiv:2408.15998}
}

@article{grant2006evolution,
title = {Evolution of character displacement in {D}arwin's finches},
author = {Grant, Peter R and Grant, B Rosemary},
year = 2006,
journal = {Science},
publisher = {American Association for the Advancement of Science},
volume = 313,
number = 5784,
pages = {224--226}
}

@book{losos2011lizards,
title = {Lizards in an evolutionary tree: ecology and adaptive radiation of anoles},
author = {Losos, Jonathan B},
year = 2011,
publisher = {Univ of California Press},
volume = 10
}

@article{brawand2014genomic,
title = {The genomic substrate for adaptive radiation in African cichlid fish},
author = {Brawand, David and Wagner, Catherine E and Li, Yang I and Malinsky, Milan and Keller, Irene and Fan, Shaohua and Simakov, Oleg and Ng, Alvin Y and Lim, Zhi Wei and Bezault, Etienne and others},
year = 2014,
journal = {Nature},
publisher = {Nature Publishing Group UK London},
volume = 513,
number = 7518,
pages = {375--381}
}

@techreport{wah2011cub,
title = {Caltech-UCSD Birds-200-2011},
author = {Wah, C. and Branson, S. and Welinder, P. and Perona, P. and Belongie, S.},
year = 2011,
number = {Cns-tr-2011-001},
institution = {California Institute of Technology}
}

@article{russakovsky2015imagenet,
title = {Imagenet large scale visual recognition challenge},
author = {Russakovsky, Olga and Deng, Jia and Su, Hao and Krause, Jonathan and Satheesh, Sanjeev and Ma, Sean and Huang, Zhiheng and Karpathy, Andrej and Khosla, Aditya and Bernstein, Michael and others},
year = 2015,
journal = {International journal of computer vision},
publisher = {Springer},
volume = 115,
pages = {211--252}
}
}
%\balance

\clearpage

\section*{Appendices}

We provide additional details omitted in the main text:
\begin{enumerate}[nosep]
    \item{\cref{app:limitations}}: Limitations
    \item{\cref{app:training-details}: SAE Training Details (for \cref{sec:sae-training})}
    \item{\cref{app:understanding-details}: SAE-Enabled Analysis Details (for \cref{sec:understanding})}
    \item{\cref{app:classification-details}: Classification Details (for \cref{sec:classification})}
    \item{\cref{app:semseg-details}: Semantic Segmentation Details (for \cref{sec:semseg})}
    \item{\cref{app:compute-requirements}: Compute Requirements}
    % \item{\cref{app:semseg-quantitative-details}: Quantitative Semantic Segmentation Details (for \cref{sec:semseg-quantitative})}
\end{enumerate}

\section{Limitations}\label{app:limitations}
    
\paragraph{Manual, potentially selective feature mining.}
Discovering interesting SAE units still depends on human browsing of exemplar grids.  
This process is slow, may overlook important concepts, and risks cherry-picking—an issue flagged by reviewers.  
Automated search and concept-label assignment (e.g.\ via language supervision) remain important future directions.

\paragraph{Qualitative evidence only.}
Our work relies significantly on qualitative evaluation through visual inspection of feature activations and manipulation effects. 
While this approach provides intuitive validation of feature interpretations, it lacks standardized quantitative metrics for measuring both interpretability and control effectiveness. 
Developing robust metrics that quantify semantic alignment, manipulation specificity, and cross-model comparability would advance this research direction from exploratory investigation toward systematic evaluation. 
Such metrics would facilitate comparing different interpretability methods and tracking progress in the field.

\paragraph{Causality caveats and distribution shift.}
Latent edits approximate knockout experiments, yet large interventions can push activations off-manifold.  
Feature correlations and non-linear interactions complicate clean causal attribution, limiting the strength of causal conclusions.

\paragraph{Coverage and stability of the concept dictionary.}
SAEs are not guaranteed to surface every concept a model encodes \citep{fel2025archetypal}.  
Low-activation units can be polysemantic and may vary across random initialisations.  
Systematic studies of exhaustiveness and retraining stability are deferred to future work.

\paragraph{Task scope and baseline overlap.}
We validate only on fine-grained classification and semantic segmentation.  
Other tasks (e.g.\ bias removal, captioning) and prior frameworks such as DN-CBM already enable editable concepts for classification; broader task coverage and explicit baselines would clarify added value.

\paragraph{Pitfalls of quantitative faithfulness metrics.}
Prior work shows that standard saliency “faithfulness scores’’ can be misleading under distribution shift \citep{hooker2019benchmark,adebayo2018sanity}.  
Following the qualitative-research guidance of \citet{olah2024qualitative}, we prioritise human-falsifiable dashboards until community-accepted metrics emerge.

\section{SAE Training Details}\label{app:training-details}

Our training code is \href{https://github.com/anonymous-saev/saev/blob/main/saev/training.py}{publicly available}, along with instructions to \href{https://github.com/anonymous-saev/saev/blob/main/saev/guide.md}{train your own SAEs for ViTs}.
Below, we describe the technical details necessary to re-implement our work.

We save all patch-level activation vectors for a given dataset from layer $l$ of a pre-trained ViT to disk, stored in a sharded format in 32-bit floating points.
Future work should explore lower-precision storage.
In practice, we explore $12$-layer ViTs and record activations from layer $11$ after all normalization.

We then train a newly initialized SAE on \num{100}M activations.
$W_\text{enc}$ and $W_\text{dec}$ are initialized using PyTorch's \texttt{kaiming\_uniform\_} initialization and $b_\text{enc}$ is zero-initialized.
$b_\text{dec}$ is initialized as the mean of \num{524288} random samples from the dataset, as recommended by prior work \citep{templeton2024scaling,templeton2024update}.

We use mean squared error $||\mathbf{\hat{x}} - \mathbf{x}||_2^2$ as our reconstruction loss, L1 length of $f(\mathbf{x})$ scaled by the current sparsity coefficient $\lambda$ as our sparsity loss, and track L0 throughout training.

The learning rate $\eta$ and sparsity coefficient $\lambda$ are linearly scaled from $0$ to their maximum over 500 steps each and remain at their maximum for the duration of training.

Columns of $W_\text{dec}$ are normalized to unit length after every gradient update, and gradients parallel to the columns are removed before gradient updates.

\begin{table}[t]
    \centering
    \small
    \caption{Hyperparameters for training SAEs. Sparsity coefficient $\lambda$ and learning rate $\eta$ are chosen qualitatively by inspecting discovered features.}
    \label{tab:hyperparameters}
    \begin{tabular}{lr}
        \toprule
        Hyperparameter & Value \\ 
        \midrule
        Hidden Width & \num{24576} ($32\times$ expansion) \\
        Sparsity Coefficient $\lambda$ & $\{4\times10^{-4},8\times10^{-4},1.6\times10^{-3}\}$ \\
        Sparsity Coefficient Warmup & \num{500} steps \\
        Batch Size & \num{16384} \\
        Learning Rate $\eta$ & $\{3\times10^{-4},1\times10^{-3},3\times10^{-3}\}    $ \\
        Learning Rate Warmup & \num{500} steps \\
        \bottomrule
    \end{tabular}
\end{table}

\section{SAE-Enabled Analysis Details}\label{app:understanding-details}

For our comparative analysis of CLIP and DINOv2 using sparse autoencoders (SAEs), we employ a qualitative discovery approach focused on identifying meaningful semantic features learned by each model. 
This section details our experimental methodology.

\paragraph{Model and Layer Selection:}
We use pretrained CLIP ViT-B/16 \citep{radford2021learning} and DINOv2 ViT-B/14 \citep{oquab2023dinov2} models as our base vision transformers. 
For both models, we extract activations from the \num{11}th transformer layer (second-to-last layer), as features at this depth capture high-level semantic information while not being overly specialized to the model's specific pretraining objective.

\paragraph{SAE Architecture and Training:}
Following the procedure detailed in \cref{sec:methodology}, we train identical SAE architectures for both models.
We use the same preprocessing and normalization steps as described in the main methodology section, ensuring consistency across experiments.

\paragraph{Feature Discovery Process:}
Our feature discovery process uses a manual exploration approach:
\begin{enumerate}
    \item We build a custom UI for feature visualization that displays exemplar images that maximally activate SAE features.
    \item Features are sorted by two criteria: sparsity (lower is better) and maximum activation value (higher is better).
    \item We systematically examine hundreds of features, recording observations about semantic patterns.
    \item When we find potentially interesting features (e.g., country-specific representations or abstract concepts), we conduct further targeted investigations.
\end{enumerate}
This iterative process involves multiple hours of exploration per model to identify prominent patterns in feature representations between CLIP and DINOv2.

\paragraph{Feature Verification:}
To verify the semantic consistency of discovered features, we employ a deliberate testing approach:
\begin{enumerate}
    \item For each potentially interesting feature, we examine both positive examples (images expected to contain the concept) and negative examples (images without the concept).
    \item We deliberately select challenging negative examples that shared visual similarities but differed in the target semantic concept (e.g., national symbols from different countries, visually similar but semantically different accident scenes).
    \item For cross-model comparison, we attempt to find corresponding features in both models by:
    \begin{itemize}
        \item Starting with positive examples known to activate a specific feature in one model.
        \item Identifying features in the second model with high activation on the same examples.
        \item Examining maximally activating examples for those candidate features to verify semantic alignment.
    \end{itemize}
\end{enumerate}
% This verification process helps ensure that our observations about model differences are robust and not artifacts of cherry-picked examples.

\begin{figure*}[t]
    \centering
    \small
    \setlength{\tabcolsep}{1pt}
    \begin{tabular}{ccccc}
        \multicolumn{4}{c}{\texttt{CLIP-24K/7622}: ``United States of America''} & \textit{Not} USA \\ % \multirow{3}{*}{\includegraphics[width=24.82pt]{figures/clip-vs-dinov2/legend.png}} \\ 
        \cmidrule(lr){1-4} \cmidrule(lr){5-5}
        \includegraphics[width=0.18\textwidth]{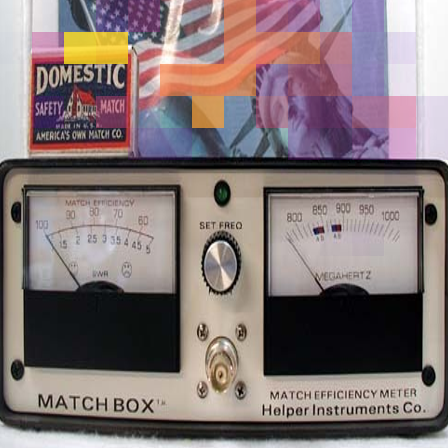} & 
        \includegraphics[width=0.18\textwidth]{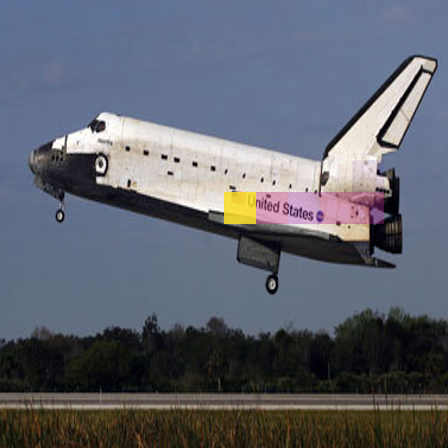} & 
        \includegraphics[width=0.18\textwidth]{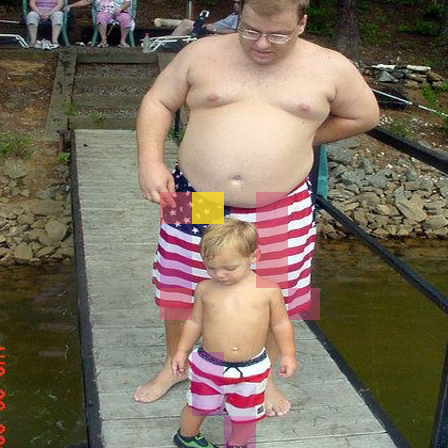} & 
        \includegraphics[width=0.18\textwidth]{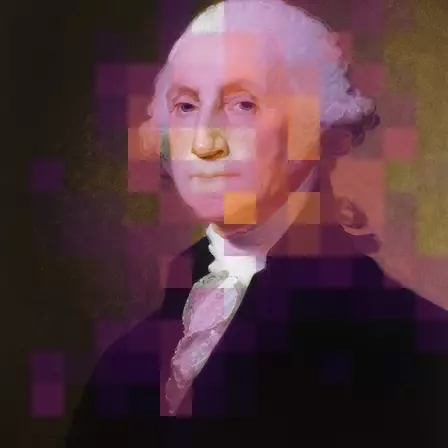} &
        \includegraphics[width=0.18\textwidth]{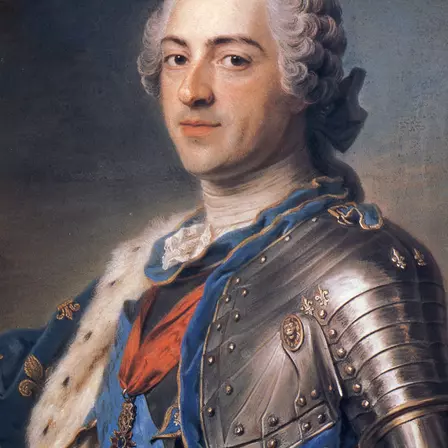} \\
        \multicolumn{4}{c}{\texttt{CLIP-24K/13871}: ``Germany''} & \textit{Not} Germany \\ % \multirow{3}{*}{\includegraphics[width=24.82pt]{figures/clip-vs-dinov2/legend.png}} \\ 
        \cmidrule(lr){1-4} \cmidrule(lr){5-5}
        \includegraphics[width=0.18\textwidth]{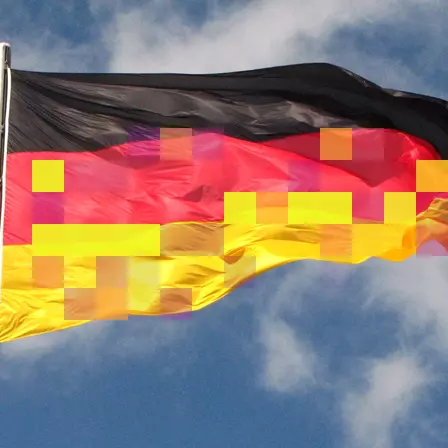} & 
        \includegraphics[width=0.18\textwidth]{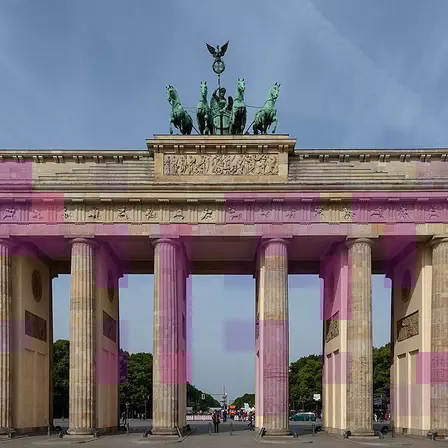} & 
        \includegraphics[width=0.18\textwidth]{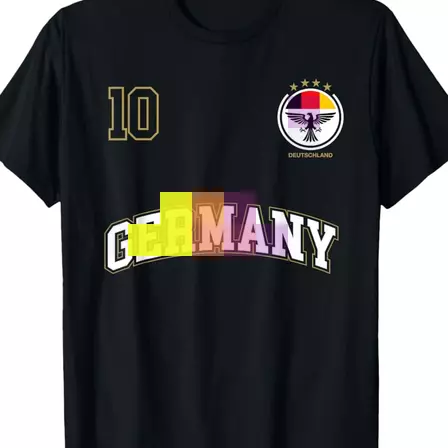} & 
        \includegraphics[width=0.18\textwidth]{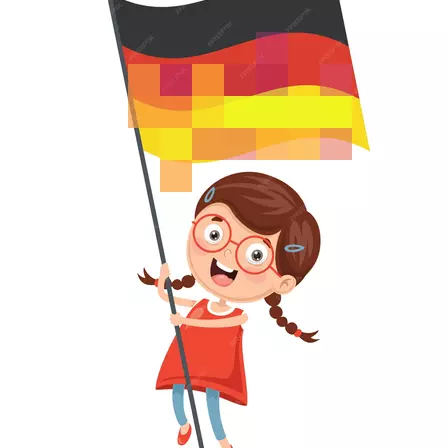} &
        \includegraphics[width=0.18\textwidth]{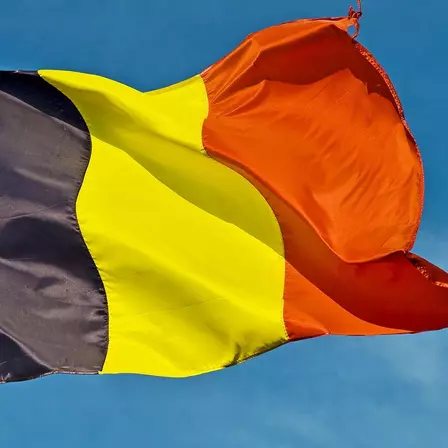}
    \end{tabular}
    \caption{
        Additional examples of cultural features learned by CLIP.
        \textbf{Top:} \texttt{CLIP-24K/7622} responds to symbolism from the United States of America, including a portrait of George Washington, but not to a portrait of King Louis XIV of France.
        \textbf{Bottom:} \texttt{CLIP-24K/13871} activates strongly on the Brandenburg Gate and other German symbols, but not on visually similar flags like the Belgian flag.}\label{fig:additional-clip-vs-dinov2-cultural}
\end{figure*}

\paragraph{Exemplar Selection for Figures:}
We manually select exemplar images in figures to effectively demonstrate the semantic properties of each feature. 
We choose positive examples to illustrate the breadth of a concept (e.g., different manifestations of ``Brazil'' or ``accidents''), and negative examples to demonstrate feature specificity (e.g., other South American imagery that does not activate the ``Brazil'' feature).
Additional examples of CLIP learning cultural features are in \cref{fig:additional-clip-vs-dinov2-cultural}.

\section{Classification Details}\label{app:classification-details}

% Detailed instructions for reproducing our results using our public codebase are \href{https://osu-nlp-group.github.io/SAE-V/contrib/classification}{available on the web}.
We use an SAE trained on \num{100}M CLIP ViT-B/\num{16} activations from layer \num{11} on ImageNet-1K's training split of \num{1.2}M images following the procedure in \cref{app:training-details}.

\subsection{Task-Specific Decoder Training (Image Classification)}

For image classification, we train a linear probe on the [CLS] token representations extracted from a CLIP-pretrained ViT-B/16 model \citep{radford2021learning} using the CUB-2011 dataset \citep{wah2011cub}. The following details describe our experimental setup to ensure exact reproducibility.

\paragraph{Data Preprocessing.}
Each input image is first resized so that its shorter side is 256 pixels, followed by a center crop to obtain a 224$\times$224 image. The resulting image is then normalized using the standard ImageNet mean and standard deviation for RGB channels. No additional data augmentation is applied.

\paragraph{Feature Extraction.}
We use a CLIP-pretrained ViT-B/16 model. From the final layer of the ViT, we extract the [CLS] token representation. The backbone is kept frozen during training, and no modifications are made to the extracted features.

\paragraph{Classification Head.}
The classification head is a linear layer mapping the [CLS] token’s feature vector to logits corresponding to the 200 classes in the CUB-2011 dataset.

\paragraph{Training Details.}
We train the linear probe on the CUB-2011 training set using the AdamW \citep{loshchilov2019adamw} optimizer for 20 epochs with a batch size of 512. 
We performed hyperparameter sweeps over the learning rate with values in \{$10^{-5}$, $10^{-4}$, $10^{-3}$, $10^{-2}$, $10^{-1}$, $1.0$\} and over weight decay with values in \{$0.1$, $0.3$, $1.0$, $3.0$, $10.0$\}. 
Based on validation accuracy, we selected a learning rate of $10^{-3}$ and a weight decay of 0.1. The CLIP backbone remains frozen during this process.
The trained linear probe achieves a final accuracy of \num{79.9}\% on the CUB-2011 validation set.

\subsection{Additional Examples}

\begin{figure*}
    \centering
    \small
    \includegraphics[width=\linewidth]{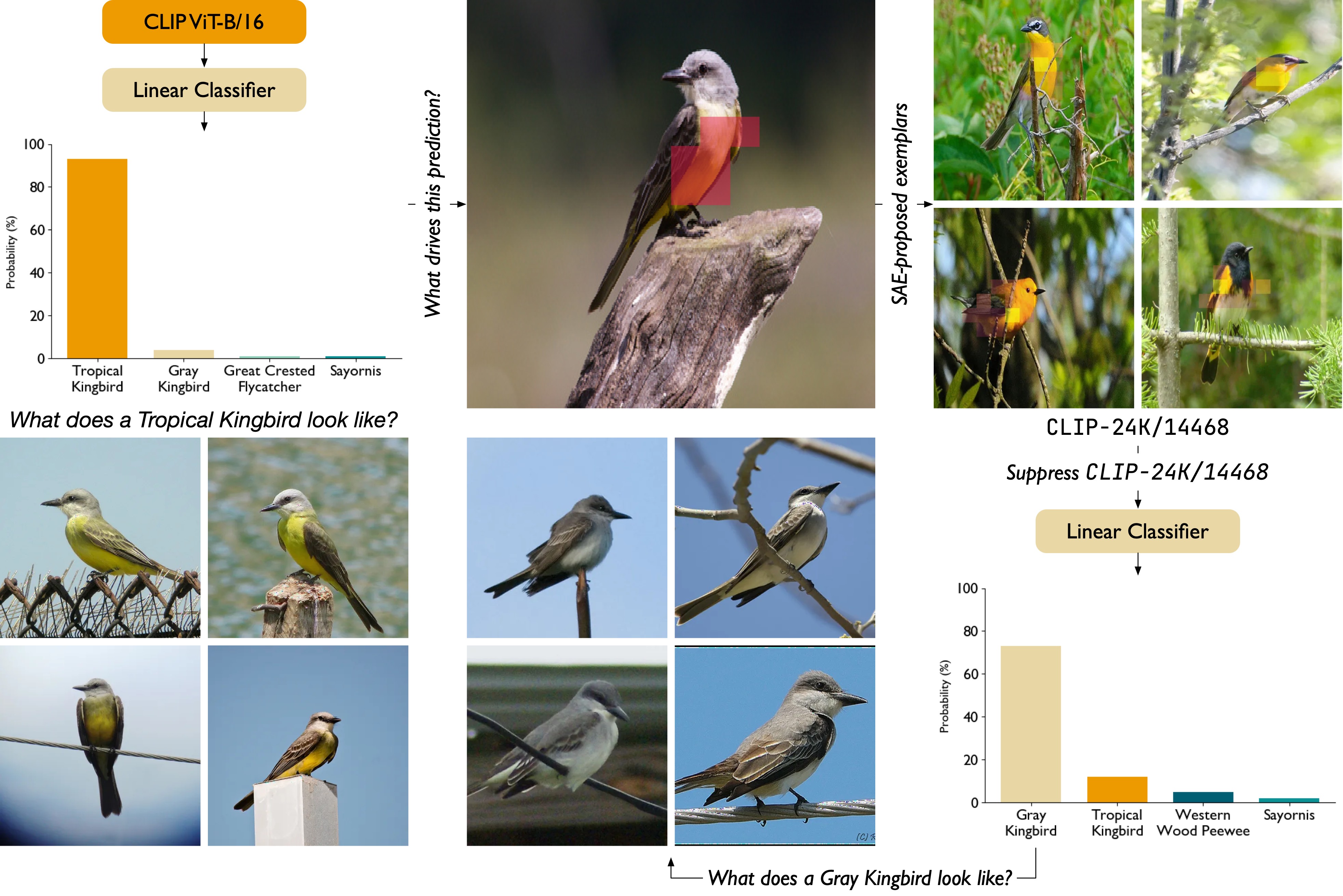}
    \caption{Tropical Kingbirds have a distinctive yellow chest. When we suppress a ``yellow feathers'' feature, our linear classifier predicts Gray Kingbird, a similar species but with a gray chest. This example is available at \href{https://anonymous-saev.github.io/saev/demos/classification?example=5099}{\url{https://anonymous-saev.github.io/saev/demos/classification?example=5099}}}\label{fig:classification-extra1}
\end{figure*}

\begin{figure*}
    \centering
    \small
    \includegraphics[width=\linewidth]{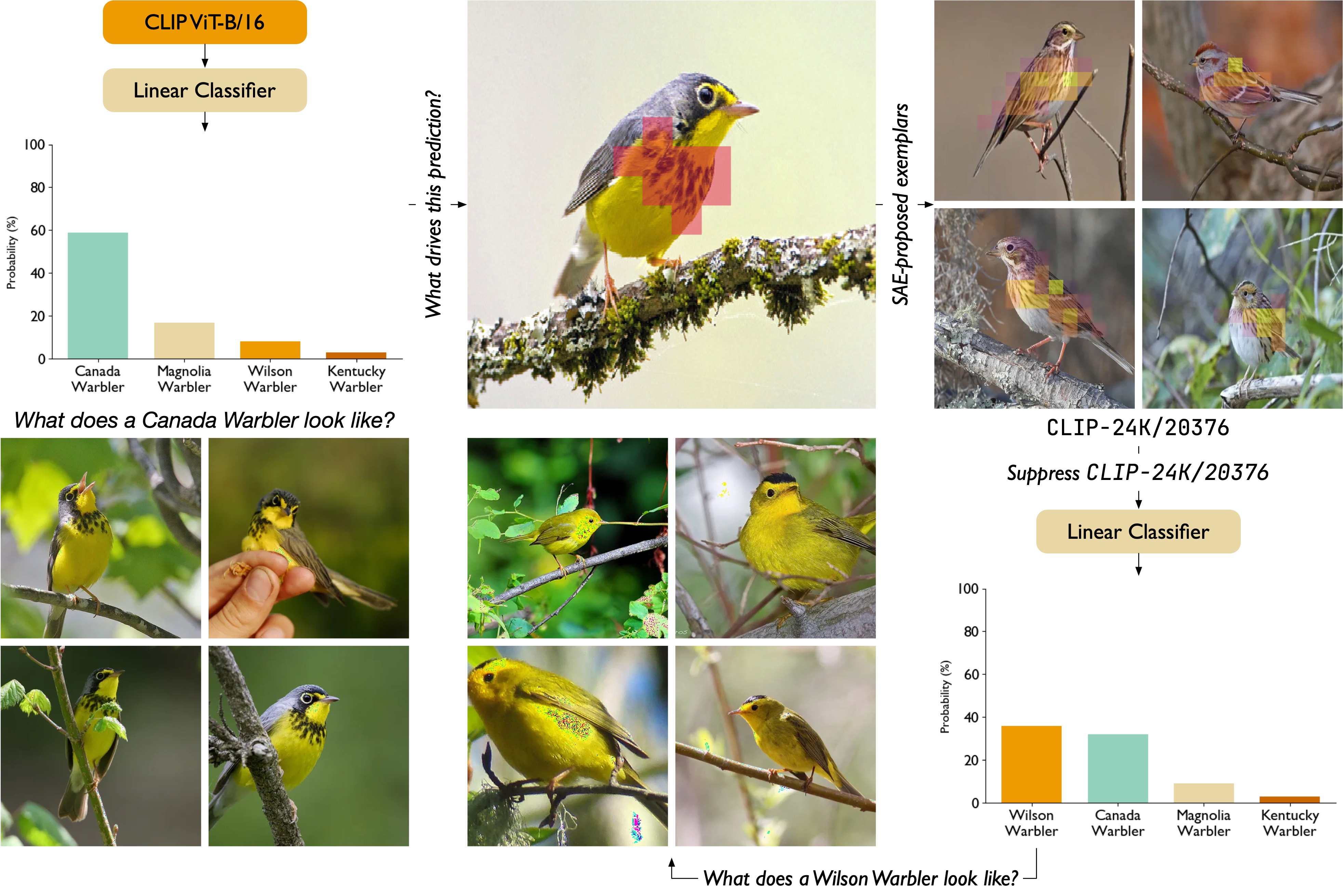}
    \caption{Canada Warblers have a distinctive black necklace on the chest. \texttt{CLIP-24K/20376} fires on similar patterns; when we suppress this feature, the linear classifier predicts Wilson Warbler, a similar species without the distinctive black necklace. This example is available at \href{https://anonymous-saev.github.io/saev/demos/classification?example=1129}{\url{https://anonymous-saev.github.io/saev/demos/classification?example=1129}}}\label{fig:classification-extra2}
\end{figure*}

\begin{figure*}
    \centering
    \small
    \includegraphics[width=\linewidth]{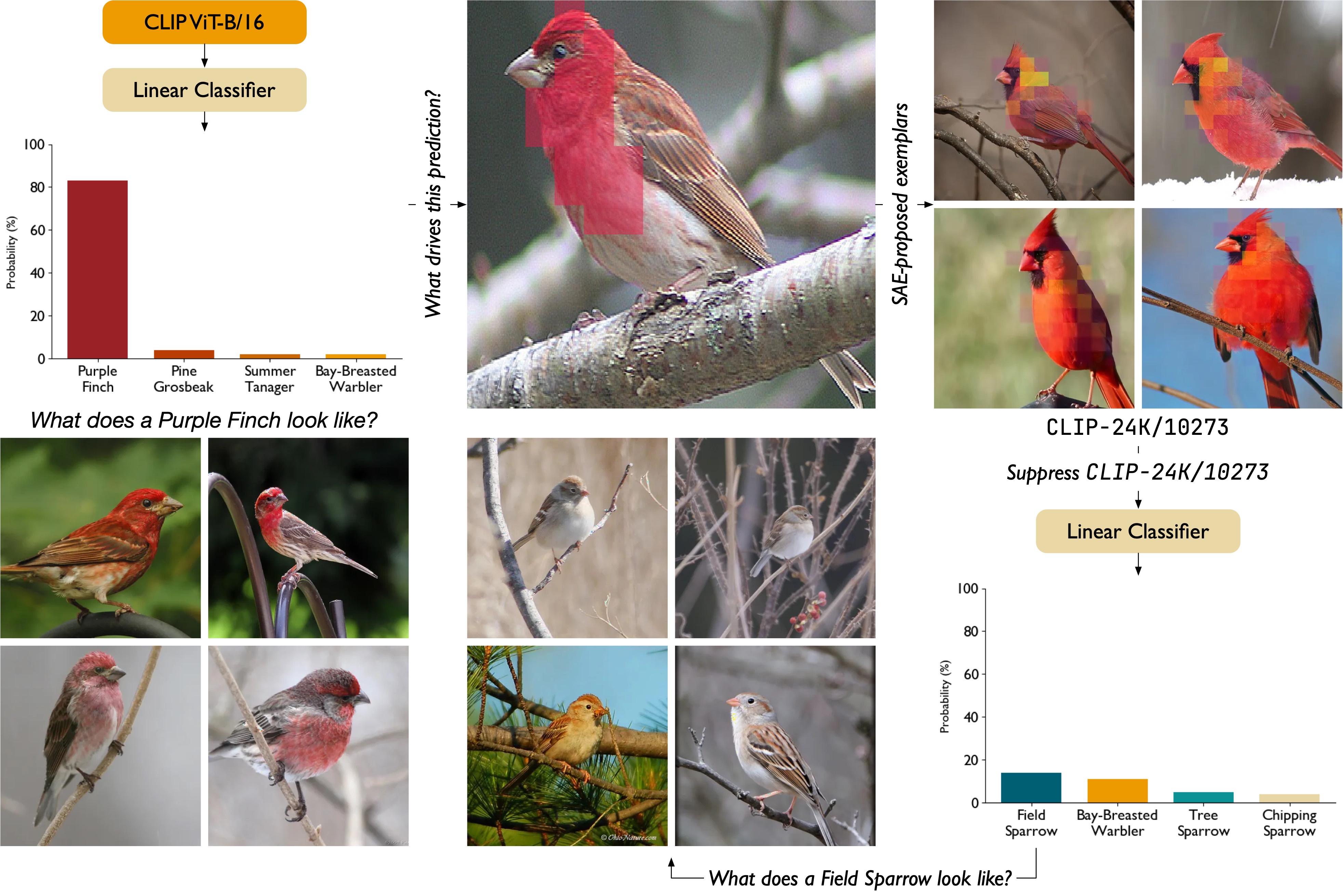}
    \caption{Purple finches have bright red coloration on the head and neck area; when we suppress \texttt{CLIP-24K/10273}, which appears to be a ``red feathers'' feature, our classifier predicts Field Sparrow, which has similar wing banding but no red coloration. 
    This example is available at \href{https://anonymous-saev.github.io/saev/demos/classification?example=4139}{\url{https://anonymous-saev.github.io/saev/demos/classification?example=4139}}}\label{fig:classification-extra3}
\end{figure*}

We provide additional examples of observing classification predictions, using SAEs to form a hypothesis explaining model behavior, and experimentally validating said hypothesis through feature suppression in \cref{fig:classification-extra1,fig:classification-extra2,fig:classification-extra3}.
Each figure links to a live, web-based demo where readers can complete the ``observe, hypothesize, experiment'' sequence themselves.

% \begin{figure*}
%     \centering
%     \small
%     \includegraphics[width=\linewidth]{figures/classification-extra4.pdf}
%     \caption{\href{https://anonymous-saev.github.io/saev/demos/classification?example=5099}{\url{https://anonymous-saev.github.io/saev/demos/classification?example=5269}}}\label{fig:classification-extra4}
% \end{figure*}

\section{Semantic Segmentation Details}\label{app:semseg-details}

Detailed instructions for reproducing our results using our public codebase are \href{https://github.com/anonymous-saev/saev/blob/main/contrib/semseg/reproduce.md}{available on the web}.

\subsection{Task-Specific Decoder Training (Semantic Segmentation)}

For semantic segmentation, we train a single linear segmentation head on features extracted from a frozen DINOv2-pretrained ViT-B/14 model \citep{oquab2023dinov2} using the ADE20K dataset \citep{zhou2017ade20k}. 
We aim to map patch-level features to 150 semantic class logits. 
The following details describe our experimental setup to ensure exact reproducibility.

\paragraph{Data Preprocessing.}
Each image is first resized so that its shorter side is 256 pixels, followed by a center crop to obtain a 224$\times$224 image. The cropped images are normalized using the standard ImageNet mean and standard deviation for RGB channels. No additional data augmentation is applied.

\paragraph{Feature Extraction.}
We use a DINOv2-pretrained ViT-B/14 with 224$\times$224 images, which results in a $16\times16$ grid of $14\times14$ pixel patches. 
The final outputs from the ViT (after all normalization layers) are used as features. 
We exclude the [CLS] token and any register tokens, retaining only the patch tokens, thereby producing a feature tensor of shape 1$6\times16\times d$, where $d=768$ is the ViT's output feature dimension.

\paragraph{Segmentation Head.}
The segmentation head is a linear layer applied independently to each patch token. 
This layer maps the $d$-dimensional feature vector to a \num{150}-dimensional logit vector, corresponding to the \num{150} semantic classes. 
For visualization purposes, we perform a simple upsampling by replicating each patch prediction to cover the corresponding $14\times14$ pixel block. 
For quantitative evaluation (computing mIoU), the $16\times16$ logit map is bilinearly interpolated to the full image resolution.

\paragraph{Training Details.}
We train the segmentation head using the standard cross entropy loss. 
The DINOv2 ViT-B/14 backbone remains frozen during training, so that only the segmentation head is updated. 
We use AdamW \citep{loshchilov2019adamw} with a batch size of \num{1024} over \num{400} epochs. 
We performed hyperparameter sweeps over the learning rate with values in \{$3\times10^{-5}$, $1\times10^{-4}$, $3\times10^{-4}$, $1\times10^{-3}$, $3\times10^{-3}$, $1\times10^{-2}$\} and over weight decay values in \{$1\times10^{-1}$, $1\times10^{-3}$, $1\times10^{-5}$\}. No learning rate schedule is applied; the chosen learning rate is kept constant throughout training. The segmentation head is initialized with PyTorch's default initialization.

\paragraph{Evaluation.}
For evaluation, the predicted $16\times16$ logit maps are upsampled to the full image resolution using bilinear interpolation before computing the mean Intersection-over-Union (mIoU) with the ground-truth segmentation masks. Our final segmentation head achieves an mIoU of \num{35.1} on the ADE20K validation set.

\subsection{Additional Examples}

\begin{figure*}
    \centering
    \small
    \includegraphics[width=\linewidth]{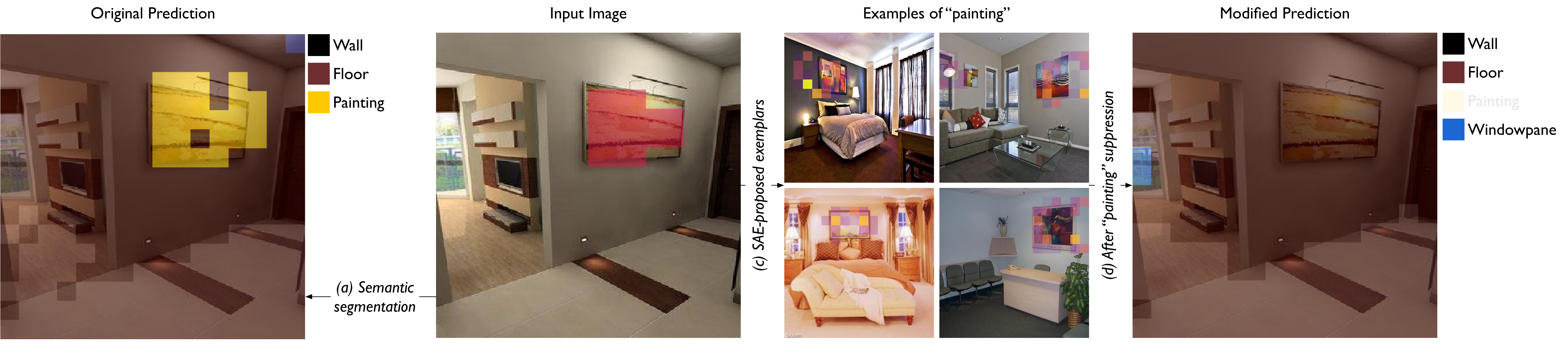}
    \caption{DINOv2 correctly identifies the framed painting. We find that \texttt{DINOv2-24K/16446} fires for paintings, and that suppressing this feature removes the painting without meaningfully affecting other parts of the image, despite modifying all patches. This example is available at \href{https://anonymous-saev.github.io/saev/demos/semseg?example=1633}{\url{https://anonymous-saev.github.io/saev/demos/semseg?example=1633}}.}\label{fig:semseg-extra1}
\end{figure*}

\begin{figure*}
    \centering
    \small
    \includegraphics[width=\linewidth]{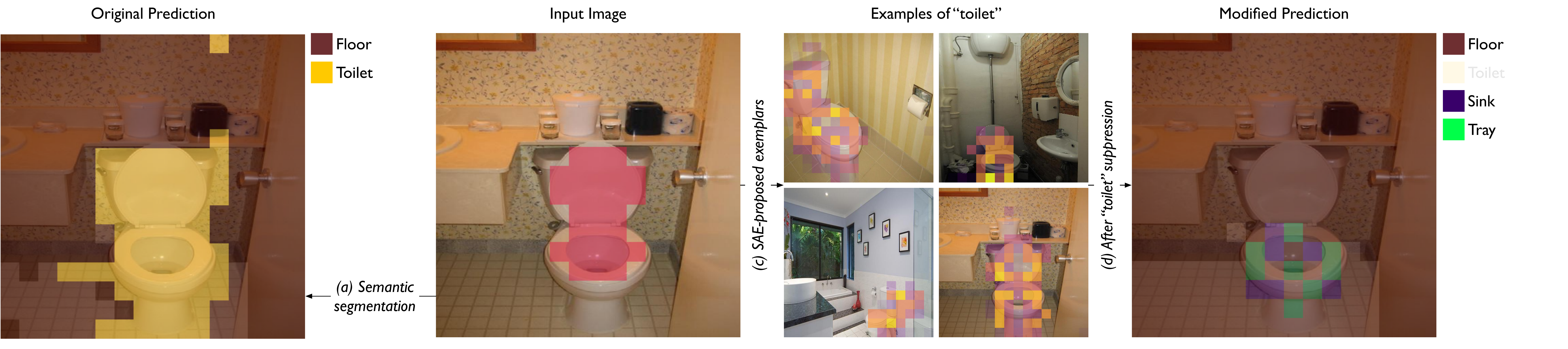}
    \caption{We find that feature \texttt{DINOv2-24K/5876} fires for toilets, and that suppressing this feature removes the toilet without meaningfully affecting other parts of the image, despite modifying all patches. This example is available at \href{https://anonymous-saev.github.io/saev/demos/semseg?example=1099}{\url{https://anonymous-saev.github.io/saev/demos/semseg?example=1099}}.}\label{fig:semseg-extra2}
\end{figure*}

\begin{figure*}
    \centering
    \small
    \includegraphics[width=\linewidth]{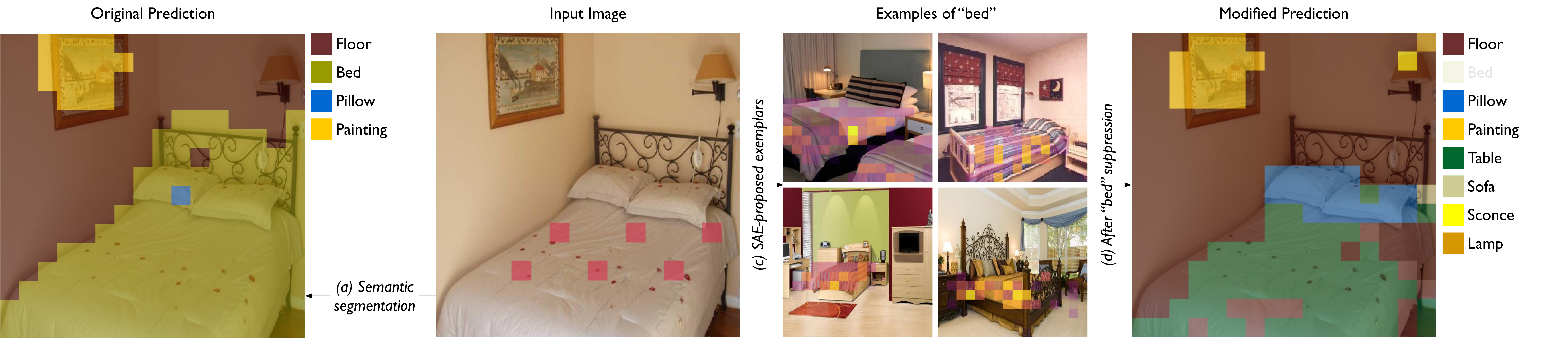}
    \caption{After suppressing a ``bed'' feature (\texttt{DINOv2-24K/18834}), the segmentation head predicts ``pillow'' for the pillows and ``table'' for the bed spread. This example is available at \href{https://anonymous-saev.github.io/saev/demos/semseg?example=1117}{\url{https://anonymous-saev.github.io/saev/demos/semseg?example=1117}}.}\label{fig:semseg-extra3}
\end{figure*}

\begin{figure*}
    \centering
    \small
    \includegraphics[width=\linewidth]{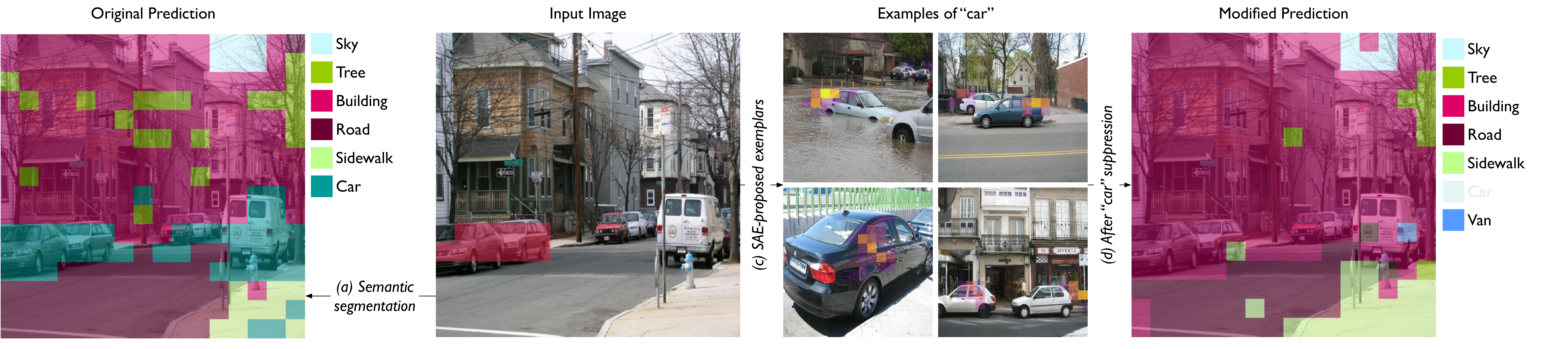}
    \caption{We remove all cars from the scene by suppressing a ``car''-like feature (\texttt{DINOv2-24K/7235}). Notably, the van on the right of the image is also removed. This example is available at \href{https://anonymous-saev.github.io/saev/demos/semseg?example=1852}{\url{https://anonymous-saev.github.io/saev/demos/semseg?example=1852}}.}\label{fig:semseg-extra4}
\end{figure*}

We provide additional examples of observing segmentation predictions, using SAEs to form a hypothesis explaining model behavior, and experimentally validating said hypothesis through feature suppression in \cref{fig:semseg-extra1,fig:semseg-extra2,fig:semseg-extra3,fig:semseg-extra4}.
These additional examples further support the idea of SAEs discovering features that are pseduo-orthogonal.
Each figure links to a live, web-based demo where readers can complete the ``observe, hypothesize, experiment'' sequence themselves.

\FloatBarrier

\section{Compute Requirements}\label{app:compute-requirements}

We use a single NVIDIA A6000 48GB GPU for all our experiments.

Our experiments have several resource-intensive steps:
\begin{enumerate}
    \item{Record ViT activations on a large image dataset. This requires running ViT inference on many images and saving the activations to disk. For a ViT-B/16 on ImageNet-1K, this equates to $1.2$M images $\times$ $196$ patches/image $\times$ $768$ floats/patch $\times$ $4$ bytes/float = $722.5$GB. For a ViT-B/14 (DINOv2), the only difference is $256$ patches/image instead of $196$, for a total of $943.7$GB. This typically takes 6-12 hours depending on disk write speed.}
    \item{Train SAEs on cached activations. This requires both good disk read speed and a GPU for acclerated training. Because SAEs are relatively small, our code enables parallel training of multiple SAEs at once to amortize the disk reads. We train 9 $32$K-dimensional SAEs on a single A6000 GPU for $100$M patches in about 8 hours.}
    \item{Save exemplar images for SAE features. This requires a GPU to run SAE inference on the training data and calculate the top-$k$ patches per feature and more disk space to save exemplar images. Saving 400 features with $k=128$ takes about 12 hours.}
\end{enumerate}

\end{document}